\documentclass[final, 5p,authoryear]{elsarticle}
\usepackage[utf8]{inputenc}

\usepackage{hyperref}

\usepackage{multirow}
\usepackage{diagbox}


\usepackage{graphicx}
\usepackage{amsmath}
\usepackage{amssymb} 
\usepackage{arydshln} 

\DeclareMathOperator*{\argmin}{\arg\!\min}

\usepackage{placeins}
\usepackage{tabu}
\usepackage{multirow}
\usepackage[normalem]{ulem} 
\usepackage{bm} 
\usepackage{color}
\usepackage{tabularx}

\newcolumntype{C}[1]{>{\centering\arraybackslash}p{#1}}
\usepackage[caption=false,font=footnotesize]{subfig}
\usepackage[perpage]{footmisc}

\makeatletter
\renewcommand*\env@matrix[1][\arraystretch]{%
	\edef\arraystretch{#1}%
	\hskip -\arraycolsep
	\let\@ifnextchar\new@ifnextchar
	\array{*\c@MaxMatrixCols c}}
\makeatother

\newcommand\Tstrut{\rule{0pt}{2ex}}

\journal{Medical Image Analysis}

\biboptions{authoryear,sort&compress} 

\begin{document}

\begin{frontmatter}

\title{A Deep Learning Framework for Unsupervised\\ Affine and Deformable Image Registration}

\author{Bob D. de Vos\fnref{fn:isi}}
\author{Floris F. Berendsen\fnref{fn:lkeb}}
\author{Max A. Viergever\fnref{fn:isi}}
\author{Hessam Sokooti\fnref{fn:lkeb}}
\author{Marius Staring\fnref{fn:lkeb}}
\author{Ivana I\v{s}gum\fnref{fn:isi}}

\fntext[fn:isi]{Image Sciences Institute, University Medical Center Utrecht and Utrecht University, Utrecht, the Netherlands}
\fntext[fn:lkeb]{Division of Image Processing of the Leiden University Medical Center, Leiden, The Netherlands}

\begin{abstract}
Image registration, the process of aligning two or more images, is the core technique of many (semi-)\allowbreak automatic medical image analysis tasks. Recent studies have shown that deep learning methods, notably convolutional neural networks (ConvNets), can be used for image registration. Thus far training of ConvNets for registration was supervised using predefined example registrations. However, obtaining example registrations is not trivial. To circumvent the need for predefined examples, and thereby to increase convenience of training ConvNets for image registration, we propose the Deep Learning Image Registration (DLIR) framework for \textit{unsupervised} affine and deformable image registration. In the DLIR framework ConvNets are trained for image registration by exploiting image similarity analogous to conventional intensity-based image registration. After a ConvNet has been trained with the DLIR framework, it can be used to register pairs of unseen images in one shot. We propose flexible ConvNets designs for affine image registration and for deformable image registration. By stacking multiple of these ConvNets into a larger architecture, we are able to perform coarse-to-fine image registration.
We show for registration of cardiac cine MRI and registration of chest CT that performance of the DLIR framework is comparable to conventional image registration while being several orders of magnitude faster.

\end{abstract}

\begin{keyword}
	deep learning \sep unsupervised learning \sep affine image registration \sep deformable image registration \sep cardiac cine MRI \sep chest CT
\end{keyword}

\end{frontmatter}


\section{Introduction}
Image registration is the process of aligning two or more images. It is a well-established technique in \mbox{(semi-)automatic} medical image analysis that is used to transfer information between images. Commonly used image registration approaches include intensity-based methods, and feature-based methods that use handcrafted image features \citep{sotiras2013,viergever2016}. Since recently, supervised and unsupervised deep learning techniques have been successfully employed for image registration \citep{jaderberg2015,wu2016,miao2016,liao2017,Krebs2017,cao2017,sokooti2017,yang2017,DeVos2017,eppenhof2018}.

Deep learning techniques are well suited for image registration, because they automatically learn to aggregate the information of various complexities in images that are relevant for the task at hand.
Additionally, the use of deep learning techniques potentially yields high robustness, because local optima may be of lesser concern in deep learning methods, i.e. zero gradients are often (if not always) at saddle points \citep{dauphin2014}. Moreover, deep learning methods like convolutional neural networks are highly parallelizable which makes implementation and execution on GPUs straight-forward and fast. As a consequence deep learning enhanced registration methods are exceptionally fast making them interesting for time-critical applications; e.g. for emerging image guided therapies like High Intensity Focused Ultrasound (HIFU), the MRI Linear Accelerator (MR-linac), and MRI-guided proton therapy.

Although not explicitly introduced as a method for image registration, the spatial transformer network (STN) proposed by \cite{jaderberg2015} was one of the first methods that exploited deep learning for image alignment.
The STN is designed as part of a neural network for classification. Its task is to spatially transform input images such that the classification task is simplified. Transformations might be performed using a global transformation model or a thin plate spline model. In the application of an STN, image registration is an implicit result; image alignment is not guaranteed and only performed when beneficial for the classification task at hand. STNs have been shown to aid classification of photographs of traffic signs, house numbers, and handwritten digits, but to the best of our knowledge they have not yet been used to aid classification of medical images.

In other studies deep learning methods were explicitly trained for image registration \citep{liao2017,miao2016,yang2017,sokooti2017,Krebs2017,cao2017,hu2018,hu2018a}. 
For example, convolutional neural networks (ConvNets) were trained with reinforcement learning to be agents that predicted small steps of transformations toward optimal alignment. \cite{liao2017} applied these agents for affine registration of intra-patient cone-beam CT (CBCT) to CT and \cite{Krebs2017} applied agents for deformable image registration of inter-patient prostate MRI.
Like intensity-based registration, image registration with agents is iterative. However, ConvNets can also be used to register images in one shot. For example, \cite{miao2016} used a ConvNet to predict parameters in one shot for rigid registration of 2D CBCT to CT volumes. Similarly, ConvNets have been used to predict parameters of a thin plate spline model. \cite{cao2017} used thin plate splines for deformable registration of brain MRI scans and \cite{eppenhof2018} used thin plate splines for deformable registration of chest CT scans. Furthermore, in the work of \cite{sokooti2017} it has been demonstrated that a ConvNet can be used to predict a dense displacement vector field (DVF) directly, without constraining it to a transformation model. Similarly, \cite{yang2017} used a ConvNet to predict the momentum for registration with large deformation diffeomorphic metric mapping \citep{beg2005}. Recently, \cite{hu2018} presented a method that employs segmentations to train ConvNets for global and local image registration. In this method a ConvNet takes fixed and moving image pairs as its inputs and it learns to align the segmentations. This was demonstrated on global and deformable registration of ultrasound and MR images using prostate segmentation.

While the aforementioned deep learning-based registration methods show accurate registration performance, the methods are all supervised, i.e. they rely on example registrations for training or require manual segmentations, unlike conventional image registration methods that are typically unsupervised. Training examples for registration have been generated by synthesizing transformation parameters for affine image registration \citep{miao2016} and deformable image registration \citep{sokooti2017,eppenhof2018}, or require manual annotations~\citep{hu2018,hu2018a}. However, generating synthetic data may not be trivial as it is problem specific. In contrast to supervised methods, training examples can be be obtained by using conventional image registration methods \citep{liao2017,Krebs2017,cao2017,yang2017}. Alternatively, unsupervised deep learning methods could be employed.
\cite{wu2016} exploited unsupervised deep learning by employing a convolutional stacked auto-encoder (CAE) that extracted features from fixed and moving images. It improved registration with Demons \citep{vercauteren2009} and HAMMER \citep{shen2002} on three different brain MRI datasets. However, while the CAE is unsupervised, the extracted features are optimized for image reconstruction and not for image registration. Thus, there is no guarantee that the extracted features are optimal for the specific image registration task. 

Unsupervised deep learning has been used to estimate optical flow \citep{yu2016, dosovitskiy2015, ilg2017} or to estimate depth \citep{garg2016} in video sequences. Such methods are related to medical image registration, but typically address different problems. They focus on deformations among frames in video sequences. These video sequences are in 2D, contain relatively low levels of noise, have high contrast due to RGB information, and have relatively small deformations between adjacent frames. In contrast, medical images are often 3D, may contain large amounts of noise, may have relatively low contrast and aligning them typically requires larger deformations.

\begin{figure}
	\centering
	\includegraphics[width=\columnwidth]{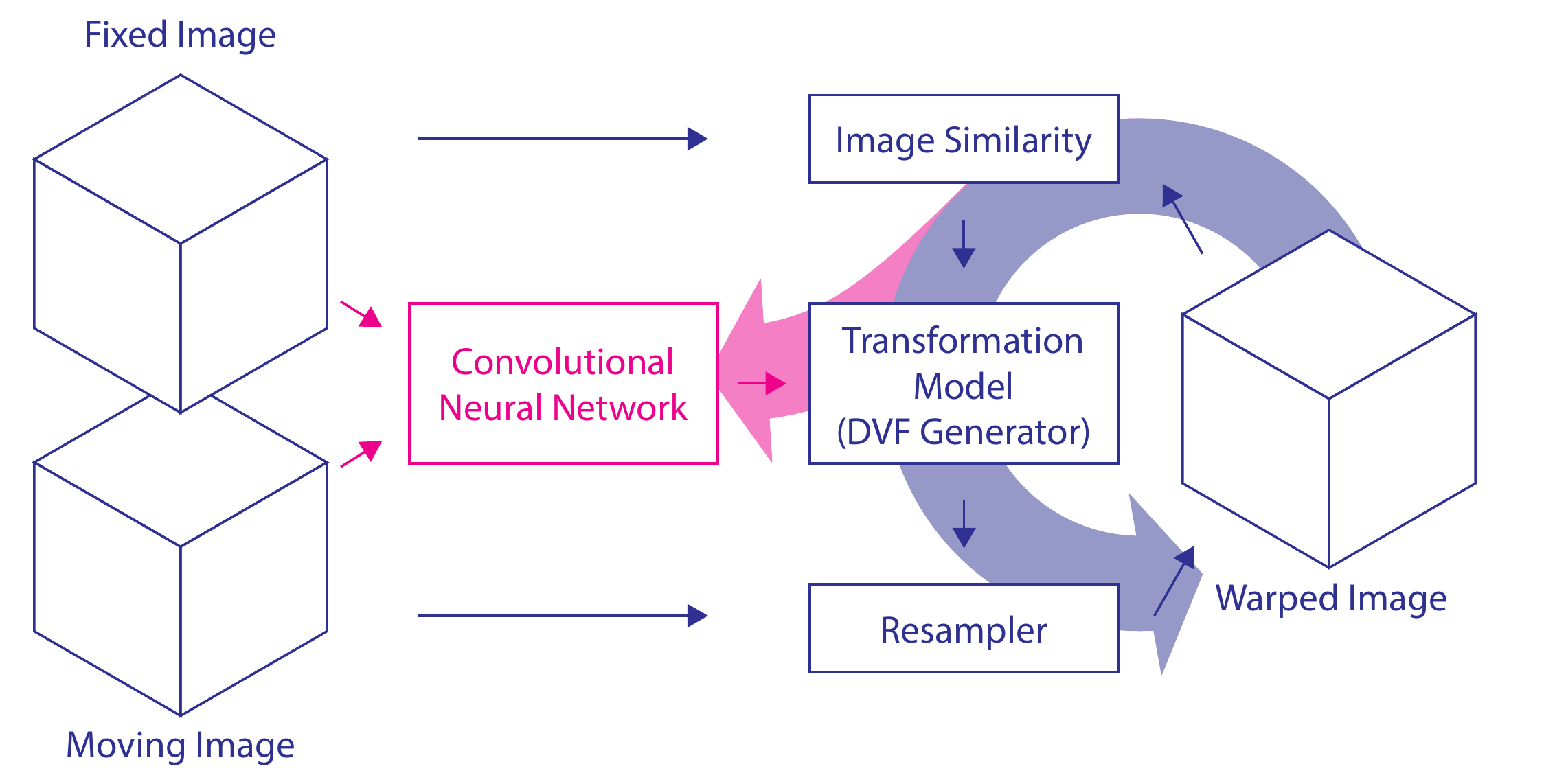}
	\caption{Schematic representation of the deep learning image registration (DLIR) framework. The DLIR training procedure is similar to a conventional iterative image registration framework (blue), but adding a ConvNet in this framework (red) allows unsupervised training for image registration. Unlike in conventional image registration, where image similarity is used to iteratively update the transform parameters directly (large blue arrow), image similarity is used in the DLIR framework to update the weights of the ConvNet by back propagation (large red arrow). Consequently, a trained ConvNet can output a transformation that aligns the input images in one shot.}
	\label{fig:framework}
\end{figure}

We propose a Deep Learning Image Registration (DLIR) framework: an unsupervised technique to train ConvNets for medical image registration tasks. In the DLIR framework, a ConvNet is trained for image registration by exploiting image similarity between fixed and moving image pairs, thereby circumventing the need for registration examples.
The DLIR framework bears similarity with a conventional iterative image registration framework, as shown in Figure~\ref{fig:framework}. 
However, in contrast to conventional image registration, the transformation parameters are not directly optimized, but indirectly, by optimizing the ConvNet's parameters.
In the DLIR framework the task of a ConvNet is to learn to predict transformation parameters by analyzing fixed and moving image pairs. The predicted transformation parameters are used to make a dense displacement vector field (DVF). The DVF is used to resample to moving image into a warped image that mimics the fixed image. During training, the ConvNet learns the underlying patterns of image registration by optimizing image similarity between the fixed and warped moving images. Once a ConvNet is trained, it has learned the image registration task and it is able to perform registration on pairs of unseen fixed and moving images in one shot, thus non-iteratively.

The current paper extends our preliminary study of unsupervised deformable image registration \citep{DeVos2017} in several ways. 
First, 
we extend the analysis from 2D to 3D images. 
Second,
we perform B-spline registration with transposed convolutions, which results in high registration speeds, reduces memory footprint, and allows simple implementation of B-spline registration on existing deep learning frameworks.
Third, borrowed from conventional image registration where regularization often is an integral part in transformation models~\citep{sotiras2013}, we include a bending energy penalty term that encourages smooth displacements.
Fourth,
we present ConvNet designs for affine as well as deformable registration.
Fifth, 
we introduce multi-stage ConvNets for registration of coarse-to-fine complexity in multiple-levels and multiple image resolutions by stacking ConvNets for affine and deformable image registration. Such a multi-stage ConvNet can perform registration tasks on fixed and moving pairs of different size, similarly to conventional iterative intensity-based registration strategies.
Finally, 
in addition to evaluation on intra-patient registration of cardiac cine MR images, we conduct experiments on a diverse set of low-dose chest CTs for inter-patient registration, and we evaluate the method on the publicly available DIR-Lab dataset for image registration~\citep{castillo2009a,castillo2009b}.

\section{Method}
\label{sec:method:dir}
In image registration the aim is to find a coordinate transformation ${\bm T:I_{F}\rightarrow I_{M}}$ that aligns a fixed image $I_{F}$ and a moving image $I_{M}$.
In conventional image registration similarity between the images is optimized by minimizing a dissimilarity metric $L$:
\begin{equation*}
{\bm\hat{\mu}} = \underset{\bm\mu}{\argmin}\{\,L(\bm T_{\bm\mu} ; I_{F}, I_{M}) + R(\bm T_{\bm\mu})\,\}\,,
\end{equation*}
where $\bm T_{\bm\mu}$ is parameterized by transformation parameters ${\bm\mu}$ and $R$ is an optional regularization term to encourage smoothness of the transformation $\bm T_{\bm\mu}$. Several dissimilarity metrics might be used, e.g. mean squared difference, normalized cross-correlation, and mutual information. Provided the metric is differentiable, optimal transformation parameters can be found by performing (stochastic) gradient descent. In the DLIR framework the ConvNet's task is to predict these transformation parameters using $I_{F}$ and $I_{M}$ as its inputs:
\begin{equation*}
{\bm\mu} = f_{\bm\theta}(I_F, I_M)\,,
\end{equation*}
where $f$ denotes the ConvNet and $\bm{\theta}$ the ConvNet's parameters. By minimizing dissimilarity $L$, a ConvNet can be trained for image registration as follows:
\begin{equation*}
\bm\hat{\theta} = \underset{{\bm\theta}}{\argmin}\{\,L(\bm T_{\bm\mu_{\bm\theta}} ; I_{F}, I_{M}) + R(\bm T_{\bm\mu_{\bm\theta}})\,\}\,.
\end{equation*}
Note that the parameters $\bm{\theta}$ of the ConvNet are optimized and not the parameters $\bm{\mu}$ of the mapping function $\bm T$. The ConvNet predicts the parameters $\bm{\mu}$ and is trained by the DLIR framework by exploiting image similarity between pairs of fixed and moving input images, i.e. image dissimilarity is calculated between the fixed and the warped moving images and is used as a loss function for ConvNet training.
While the DLIR framework could work with other transformation models, below we work out the architectures for an affine (global) parameterization and a B-spline deformable (local) parameterization of the transformation model.
In the following subsections we provide details about the DLIR framework, we describe specific ConvNet designs for affine and for deformable image registration, and finally we describe a ConvNet design consisting of multiple stages to perform multi-resolution and multi-level image registration. We refer the reader to \cite{sotiras2013} for an extensive review on image registration techniques.

\subsection{Affine Image Registration}
Affine transformation is often the first step in image registration, since it simplifies the optimization of subsequent more complex image registration steps. 
Considering that the affine transformation model is global, we designed a ConvNet that analyzes a pair of input images globally. Considering that medical images often have different image sizes, the proposed ConvNet analyzes fixed and moving images in separate pipelines. The separate pipelines analyze input images independently and therefore eliminate the need for cropping or padding of input image pairs to the same size. In each pipeline the final feature maps will be of different size. Thus global average pooling~\citep{lin2014} is applied to output one feature per feature map by taking the average of each feature-map. An additional benefit is that global pooling forces the network to encode orientations and affine transformations globally. Subsequently, the network can be connected to a neural network work that will decode the relative orientations of the fixed and moving images and convert those to 12 affine transformation parameters: three translation, three rotation, three scaling, and three shearing parameters.

\begin{figure}
	\centering
	\includegraphics[width=\columnwidth]{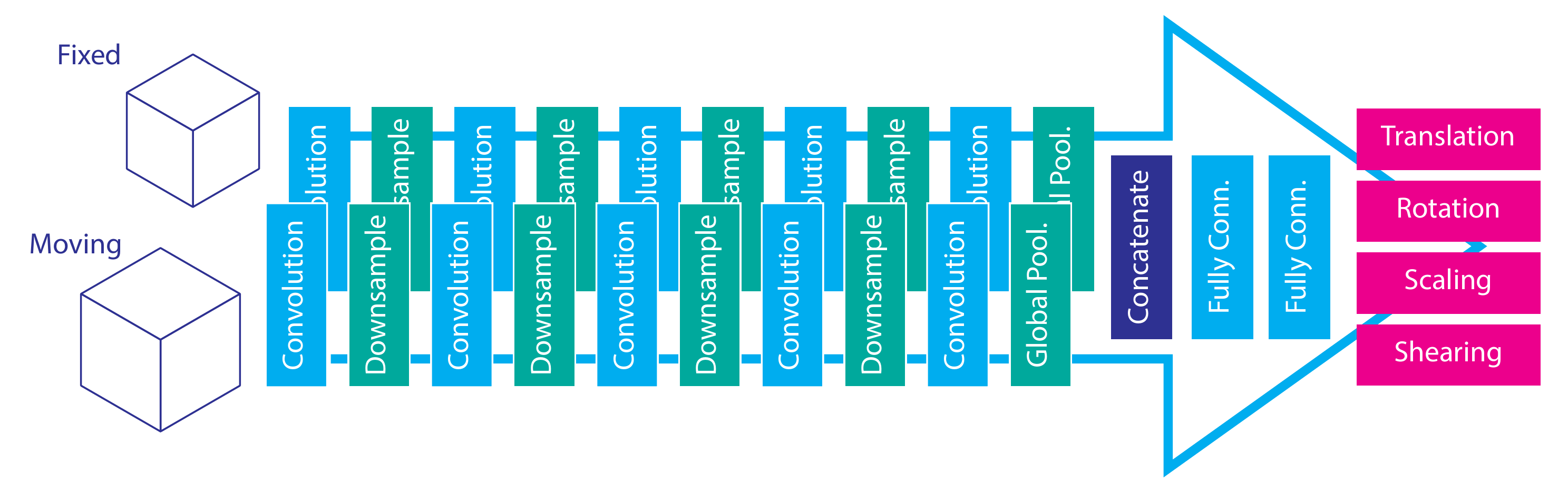}
	\caption{A ConvNet design for affine image registration. The network analyzes pairs of fixed and moving images in separate pipelines. 
	Ending each pipeline with global average pooling enables analysis of input images of different sizes, and allows concatenation with the fully connected layers that have a fixed number of nodes connected to 12 affine transformation parameter outputs.}
	\label{fig:affine}
\end{figure}

Figure~\ref{fig:affine} illustrates our ConvNet design for affine image registration. The two separate pipelines analyze input pairs of fixed and moving images and each consist of five alternating $3\times3\times3$ convolution layers and $2\times2\times2$ downsampling layers. The number of these layers may vary, depending on task complexity and input image size. The weights of the layers are shared between the two pipelines to limit the number of total parameters in the network. 

\subsection{Deformable Image Registration}
Deformable transformation models can account for local deformations that often occur in medical images.
Deformable image registration can be achieved with several transformation models. Here we opt for B-splines \citep{rueckert1999} because of their inherent smoothness and local support property: a B-spline control point only affects a specific area in an image, in contrast to e.g. a thin plate spline which has global support. In our ConvNet design we exploit this property by choosing a receptive field that overlaps the support size of the B-spline basis functions, i.e. at least four times the grid spacing for a third order B-spline kernel.
The ConvNet takes patches from fixed and moving images and predicts the B-spline control point displacements within that patch. By using a fully convolutional patch-based ConvNet design inspired by \cite{long2015}, input images of arbitrary dimensions can be analyzed efficiently.

\begin{figure}
	\centering
	\includegraphics[width=\columnwidth]{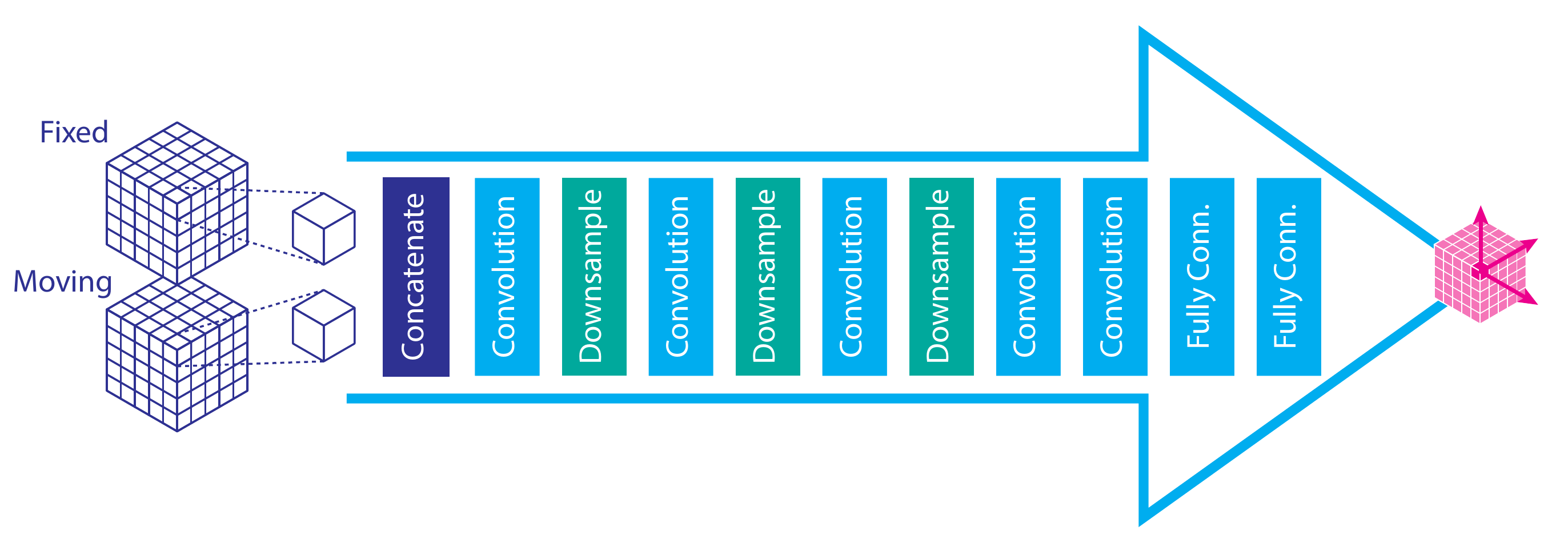}
	\caption{A patch-based ConvNet design for deformable image registration. The ConvNet takes fixed and moving image pairs of equal size as its input--e.g. pre-aligned with affine registration--and outputs a B-spline 3D displacement vector for each patch. The patch and B-spline grid dimensions determine the number of downsampling layers, thus each specific B-spline grid and image resolution requires a dedicated ConvNet design. The fully convolutional patch-based design efficiently generates a B-spline 3D displacement grid of any number of grid points depending on the input images sizes.}
	\label{fig:freeform}
\end{figure}

The proposed ConvNet design is shown in Figure~\ref{fig:freeform}. The ConvNet expects a pair of fixed and moving images of equal size that are concatenated. Depending on the registration problem, moving images might have to be pre-aligned first with e.g. affine registration. After concatenation, alternating layers of $3\times3\times3$ convolutions (with 0-padding) and $2\times2\times2$ downsampling are applied. The user-chosen B-spline grid spacing determines the amount of required downsampling. 
A larger grid spacing implies fewer control points, and thus a need for more downsampling layers; by adding more downsampling layers, the receptive field of the ConvNet simultaneously increases. 
The two additional $3\times3\times3$ convolution layers after the last downsampling layer enlarge the receptive field to the support size of the third order B-spline control points. Thereafter, two $1\times1\times1$ convolutional layers are applied, and these are connected to the final convolutional output layer with three $1\times1\times1$ kernels that predict the B-spline control points in each of the three directions. The final DVF, used for image resampling, can be generated from the estimated control points by B-spline interpolation.

B-spline interpolation was implemented efficiently by  transposed convolutions, also known as fractionally strided convolutions or deconvolutions. Transposed convolutions are the back-bone in ConvNet implementations. They are used to backpropagate loss through the convolutional layers. Due to the $2\times2\times2$ downsampling factors resulting in integer grid spacings we can use fixed precomputed B-spline kernels to efficiently upsample B-spline control points to a dense DVF. We  use a discrete B-spline kernel as the convolution kernel.

\begin{figure*}[]
	\centering
	\includegraphics[width=\textwidth]{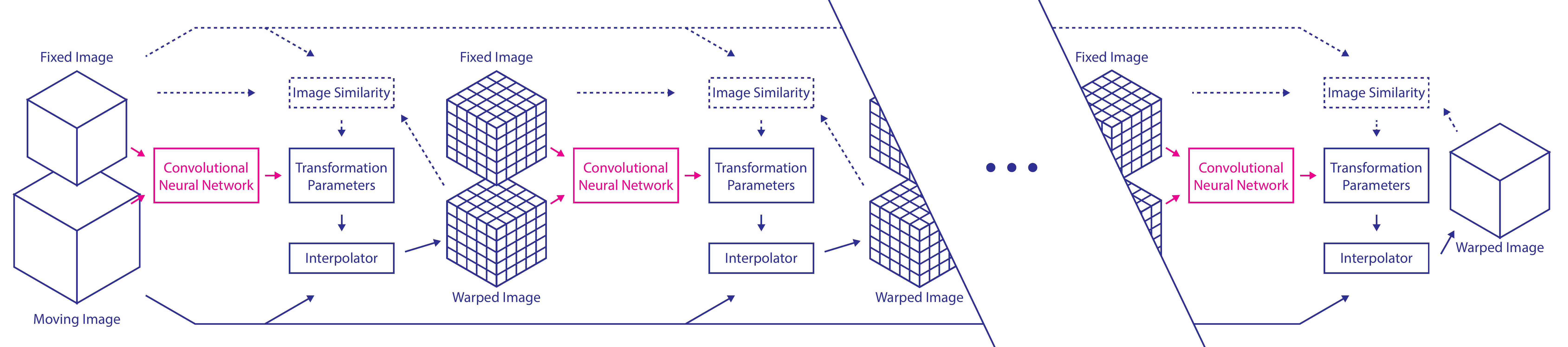}
	\caption{Schematic representation of the DLIR framework applied for hierarchical training of a multi-stage ConvNet for multi-resolution and multi-level image registration. The first stage performs affine registration of an image pair and the subsequent stages perform coarse-to-fine deformable image registration. The ConvNet in each stage is trained for its specific registration task by optimizing image similarity. The weights of the preceding ConvNets are fixed during training. This procedure prevents exploding gradients and conserves memory. Transformation parameters are passed through the network and combined at each stage to create a warped image. The warped image is passed to the subsequent stage and is used as the moving image input.}
	\label{fig:composed}
	\vspace{.2cm}
\end{figure*}

\subsection{Multi-Stage Image Registration}
Conventional image registration is often performed in multiple stages starting with affine registration, followed by coarse-to-fine stages of deformable image registration using B-splines. This hierarchical multi-stage strategy makes conventional iterative image registration less sensitive to local optima and image folding~\citep{Schnabel2001}. We adopted this strategy for the DLIR framework by stacking multiple stages of ConvNets, each with its own registration task. For example, a ConvNet for affine registration is followed by multiple ConvNets for coarse-to-fine B-spline registration, each ConvNet with a different B-spline grid spacing and images of different resolution as inputs. When multi-stage registration requires varying input resolutions, we propose average pooling (i.e. windowed averaging), which is a very common building block in deep learning frameworks.

Figure~\ref{fig:composed} illustrates how such a multi-stage ConvNet can be trained for multi-resolution and multi-level image registration. Training within the DLIR framework is performed sequentially: each stage is trained for its specific registration task, while keeping the weights of ConvNets from preceding stages fixed. After training, the multi-stage ConvNet can be applied for one-shot image registration, similar to a single ConvNet.

\subsection{Loss Function}
\label{sec:optimization}
The registration ConvNets are trained using mini-batch stochastic gradient descent, hence a differentiable loss is required. Since we perform mono-modal registration experiments, we use normalized cross correlation. Carefully chosen coarse-to-fine levels of multi-stage B-spline registration might prevent image folding and result in smooth deformations~\citep{Schnabel2001}. Alternatively, smooth deformations can be encouraged by using a bending energy penalty as proposed by \cite{rueckert1999}. The loss function we propose combines normalized cross correlation and this penalty:
\begin{equation}
    L=L_{NCC} + \alpha P\,,
\end{equation}
where $L_{NCC}$ is the negative normalized cross correlation, and $P$ the bending energy penalty with $\alpha=0$ for affine registration, and $\alpha$ empirically determined to be $0.05$ for all deformable image registration experiments. The bending energy penalty is defined as follows:
\begin{align*}
P = \frac{1}{V} \int_{0}^{X}\int_{0}^{Y}\int_{0}^{Z}\Bigg[\phantom{2}\bigg(\frac{\partial^2\textbf{T}}{\partial x^2}\bigg)^2 + \phantom{2}\bigg(\frac{\partial^2\textbf{T}}{\partial y^2}\bigg)^2 + \phantom{2}\bigg(\frac{\partial^2\textbf{T}}{\partial z^2}\bigg)^2\\
 +~2\bigg(\frac{\partial^2\textbf{T}}{\partial xy}\bigg)^2 + 2\bigg(\frac{\partial^2\textbf{T}}{\partial xz}\bigg)^2 + 2\bigg(\frac{\partial^2\textbf{T}}{\partial yz}\bigg)^2\Bigg] dx~dy~dz,
\end{align*}
where $V$ is the volume of the image domain, and \textbf{T} the local transformation. Adding this term during registration minimizes the second order derivatives of local transformations of a DVF, thereby resulting in locally affine transformations, thus enforcing global smoothness~\citep{staring2007}:

\section{Data}
Like most deep learning approaches, the DLIR framework requires large sets of training data. Publicly available datasets that are specifically provided to evaluate registration algorithms, contain insufficient training data for our approach. Therefore, we made use of large datasets of cardiac cine MRIs for intra-patient registration experiments, and low-dose chest CTs from the National Lung Screening Trial (NLST) for inter-patient registration experiments. We used manually delineated anatomical structures in these datasets for evaluation of the DLIR framework. Manually obtained delineations in the datasets were only used for final evaluation of registration performance. In addition, we used the publicly available DIR-Lab dataset. The data set is not sufficiently large to demonstrate the full potential of the proposed method, but it does provide an indication of registration performance and it enables straightforward replication of our work.

\subsection{Cardiac Cine MRI}
We included publicly available cardiac cine MRI scans from the Sunnybrook Cardiac Data \citep{Radau2009}. The data set contains 45 short-axis cine MRI images distributed over four pathology categories: healthy subjects, patients with hypertrophy, patients with heart failure and infarction, and patients with heart failure without infarction. Each scan contains 20 timepoints (i.e. volumes) encompassing the entire cardiac cycle, which results in $45\times20$ volumes in total. All scans have a slice thickness and spacing of 8\,mm and an in-plane resolution of 1.25\,mm per voxel. All scans are made with a $256\times256$ matrix and consist of about 10 slices. The data is separated into training, validation, and evaluation sets, each containing 15 scans with equally distributed pathology categories. Provided manual segmentations of left ventricle volumes at end-diastole (ED) and end-systole (ES) were used for evaluation.

\subsection{Chest CT}
\label{sec:data:chestct}
We included 2,060 chest CTs that were randomly selected from a set of scans acquired at baseline in the NLST \citep{nlst2011}. The dataset is very diverse containing scans of fourteen different CT-scanners from four vendors. 
All scans were made during inspiratory breath-hold without ECG synchronization and without contrast enhancement. Isotropic in-plane resolution of the 512$\times$512 axial slices varied between 0.45\,mm to 0.98\,mm per voxel. Slice increment ranged from 0.63\,mm to 10.0\,mm covering the thorax in 26 to 469 axial slices.
The scans were divided into 2,000 scans for training and 50 scans for validation during method development.
The remaining 10 scans provided 90 image pairs for quantitative evaluation. In each scan the entire visible aorta was delineated, including the ascending aorta, the aortic arch, and the descending aorta. In addition, ten landmarks were annotated: the carina, the aortic root, the root of the left subclavian artery, the apex of the heart, the tip of the xiphoid, the tops of the left and right lungs, the left and right sterno clavicular joints, and the tip of the spinous process of the T1 vertebra.

\subsection{DIR-Lab 4D Chest CT}
We included publicly available 4D chest CT from DIR-Lab~\citep{castillo2009a, castillo2009b}. The dataset consists of ten 4D chest CTs that encompass a full breathing cycle in ten timepoints. Isotropic in-plane resolution of 512$\times$512 axial slices ranged from 0.97\,mm to 1.98\,mm per voxel, with a slice thickness and increment of 2.5\,mm. Because the dataset is of limited size we did not separate it into seperate training, validation, and test sets. Instead, we performed leave-one-out cross-validation during evaluation.
Each scan contains 300 manually identified anatomical landmarks annotated in two timepoints, namely at maximum inspiration and maximum expiration. The landmarks serve as a reference to evaluate deformable image registration algorithms.

\section{Evaluation}
The DLIR framework was evaluated with intra-patient as well as inter-patient registration experiments. As image folding is anatomically implausible, especially in intra-patient image registration, after registration, we evaluated the topology of obtained DVFs quantitatively. For this we determined the Jacobian determinant--also known as \textit{the Jacobian}--for every point $p(i,j,k)$ in the DVF:
\begin{equation*}
det(J(i,j,k)) = 
\begin{vmatrix}[1.5]
\frac{\partial i}{\partial x} & \frac{\partial j}{\partial x} & \frac{\partial k}{\partial x} \\ 
\frac{\partial i}{\partial y} & \frac{\partial j}{\partial y} & \frac{\partial k}{\partial y} \\ 
\frac{\partial i}{\partial z} & \frac{\partial j}{\partial z} & \frac{\partial k}{\partial z} \\ 
\end{vmatrix}
\end{equation*}
A Jacobian of $1$ indicates that no volume change has occured. A Jacobian of $>1$ indicates expansion, a Jacobian between $0$ - $1$ indicates shrinkage, and a Jacobian of $\leq0$ indicates a singularity: i.e. a place where folding has occured. By indicating the fraction of foldings per image and by determining the standard deviation of the Jacobian, we can quantify the quality of the DVF.

Additionally, registration performance was evaluated using manually delineated anatomical structures and manually indicated landmarks. By propagating the delineations using obtained DVFs, registration performance can be assessed by measuring label overlap with the Dice coefficient:
$$
Dice = \frac{2|P\cap R|}{|P|+|R|},
$$
given a propagated segmentation ($P$) and a reference segmentation ($R$).

The surface distance
$$
d(x,R_S) = \min_{y\in R_S} d(x, y)),
$$
where x is a point of the propogated surface and y on the a reference surface ($R_S$), was used to calculate the average symmetric surface distance (ASD)
$$
ASD = \frac{1}{|{P_S}|+|{R_S}|}\left(\sum_{x\in P_S} d(x,R_S) + \sum_{y\in R_S} d(y,P_S) \right),
$$
where $x$ and $y$ are points on the propagated surface $P_S$ and reference surface $R_S$.
And we calculated the symmetric Hausdorff distance:
$$
HD = \max\left\{d_H(P_S,R_S), d_H(R_S,P_S)\right\},
$$
where
$$
d_H(P_S,R_S) = \max_{x\in P_S}\min_{y\in R_S} d(x, y)).
$$

For landmarks the registration error was determined as the average 3D Euclidean distance between transformed and reference points.

\section{Implementation}
\subsection{DLIR Framework}
All ConvNets were trained with the DLIR framework using the loss function provided in Section~\ref{sec:optimization}. The ConvNets were initialized with Glorot's uniform distribution \citep{glorot10a} and optimized with Adam \citep{kingma2015}.


Rectified linear units were used for activation in all ConvNets, except in the output layers. The output of the deformable ConvNets were unconstrained to enable prediction of negative B-spline displacement vectors. The outputs of affine ConvNets were constrained as follows: rotation parameters and shearing parameters were constrained between $-\pi$ and $+\pi$, the scaling parameters were constrained between $0.5$ and $1.5$, and translation parameters were unconstrained.

During training, moving images were warped using linear resampling, during evaluation segmentations were warped using nearest neighbor resampling. All experiments were performed in Python using Pytorch~\citep{paszke2017automatic} on an NVIDIA Titan-X GPU, an Intel Xeon E5-1620 3.60\,GHz CPU with 4 cores (8 threads), and 32\,GB of internal memory.

\subsection{Conventional Image Registration}
Registration performance of the DLIR framework was compared with conventional iterative intensity-based image registration using \mbox{SimpleElastix} \citep{Marstal2016}.  SimpleElastix enables integration of Elastix \citep{KleinStaring2010} in a variety of programming languages.
 
For optimal comparison, settings for conventional registration and DLIR experiments were chosen similar. Thus, similar grid settings and NCC were used. Adaptive stochastic gradient descent was used for iterative optimization. Registration stages were optimized in 500 iterations, sampling 2,000 random points per iteration. In contrast to multi-stage DLIR experiments, a Gaussian smoothing image pyramid was used in favor of windowed averaging.

\section{Intra-Patient Registration of Cardiac Cine MRI}
Intra-patient registration experiments were conducted using cardiac cine MRIs. The task was to register volumes (i.e. 3D images) within the 4D scans. 
Experiments were performed with 3-fold cross-validation. In each fold 30 images were used for training and 15 for evaluation. Given that each scan has 20 timepoints, 11,400 different permutations of image pairs were available per fold for training. Performance was evaluated using registration between images at ED and ES  by label propagation of manual left ventricle lumen segmentations. In total 90 different registration results were available for evaluation.

\subsection{ConvNet Design and Training}
To evaluate the impact of multi-stage image registration, ConvNets were trained for single stage and multi-stage deformable image registration. Initial global affine registration was not necessary, because cardiac cine MRI images only show local deformations between timepoints. Additionally, experiments were performed to study effect of the bending penalty.

Deformable registration ConvNets were designed as proposed in Section~\ref{sec:method:dir}. Downsampling was performed using average pooling. To retain information of the through-plane axis, downsampling was applied in the short-axis plane only. Experimental settings are further detailed in Table~\ref{tab:scd_settings}.

\begin{table}
	\centering
	\caption{Design of deformable image registration (DIR) stages used in single stage and multi-stage intra-patient registration of cardiac cine MRI. For multi-stage registration experiments DIR-1 and DIR-2 were sequentially applied; for single stage experiments one stage equal to DIR-2 was applied. Image resolution, grid spacing, and average number of grid points are given in x$\times$y$\times$z order.}
	\label{tab:scd_settings}
\resizebox{\columnwidth}{!}{ %
	\begin{tabular}{l|c|cc}
	     &Single Stage & \multicolumn{2}{c}{Multi-Stage} \\
		       &              DIR  &  DIR-1    & DIR-2    \\
		\hline\Tstrut
		Image resolution (mm)     &    $1.25\times1.25\times8$  & $2.50\times2.50\times16$ & $1.25\times1.25\times8$\\
		Grid spacing (mm) &           $10\times10\times8$        & $20\times20\times16$     & $10\times10\times8$\\
		Avg. grid points  & $64\times64\times10$& $32\times32\times5$ & $64\times64\times10$\\
		Mini-batch size (pairs)        &  $8$      & $16$      & $8$     \\
	\end{tabular} 
}
\end{table}

\begin{figure*}
	\centering
	\includegraphics[width=.75\textwidth]{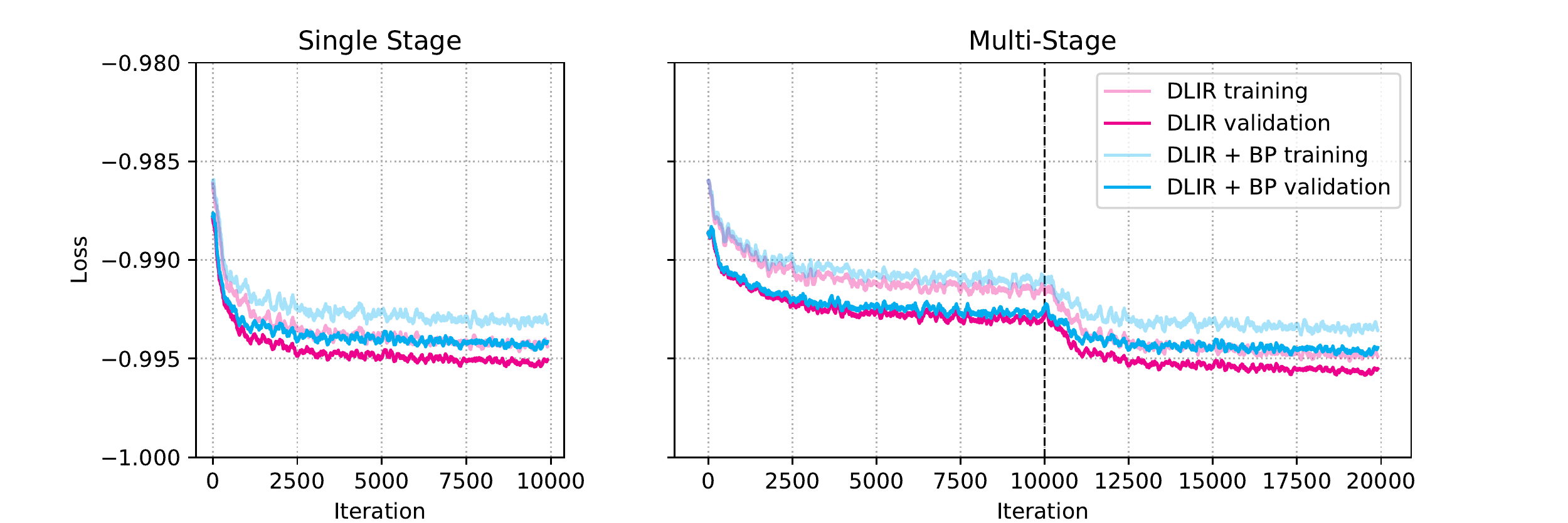}
	\caption{Learning curves showing the negative NCC during training of single stage and multi-stage ConvNets with or without bending penalty (BP) for intra-patient registration of cardiac cine MRI. Learning curves are taken from the one of the folds used in 3-fold cross validation. Augmentations were only applied to training data resulting in a relatively higher training NCC loss.}
	\label{fig:scd_learning_curves}
\end{figure*}

All ConvNets were trained with mini-batches consisting of random permutations of two timepoints taken from the same image. Prior to analysis, image intensities were linearly scaled from 0 to 1 based on the minimum intensity and 99\textsuperscript{th} percentile of the maximum intensity. During training fixed and moving image pairs were correspondingly augmented by random in-plane rotations of 90, 180, and 270 degrees and random in-plane cropping of at maximum $\pm16$ voxels. Registration stages were trained in 10,000 iterations. Each fold was trained in approximately 5 hours for single stage registration and 8 hours for multi-stage registration. Figure~\ref{fig:scd_learning_curves} shows the development of training and validation NCC between image pairs during training of one of the folds. Overfitting did not occur in the experiments, instead the training error was higher than the validation error due to the random croppings applied on the training set only.

\subsection{Results}
\begin{figure*}
\centering
		\subfloat[Fraction of folding]{\includegraphics[width = .75\textwidth]{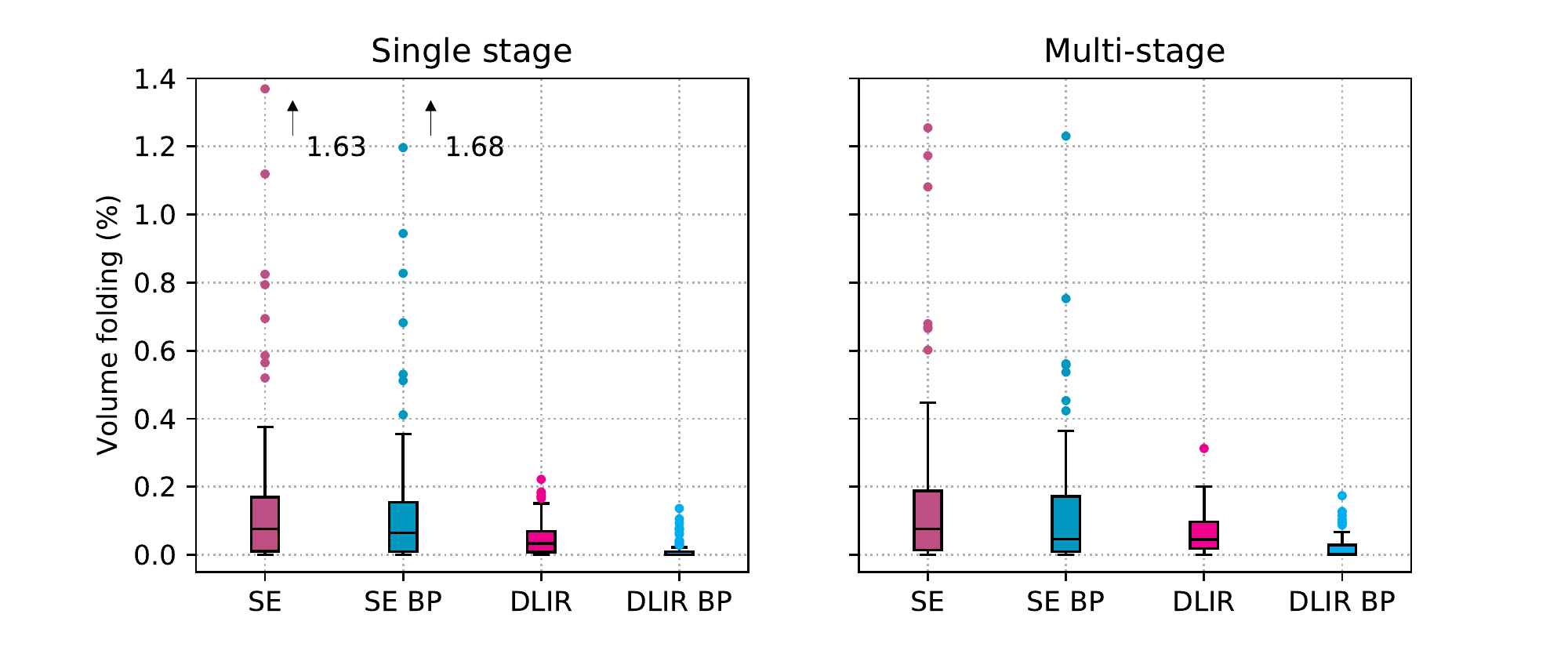}}\\
		\subfloat[Standard deviation of Jacobian determinant]{\includegraphics[width = .75\textwidth]{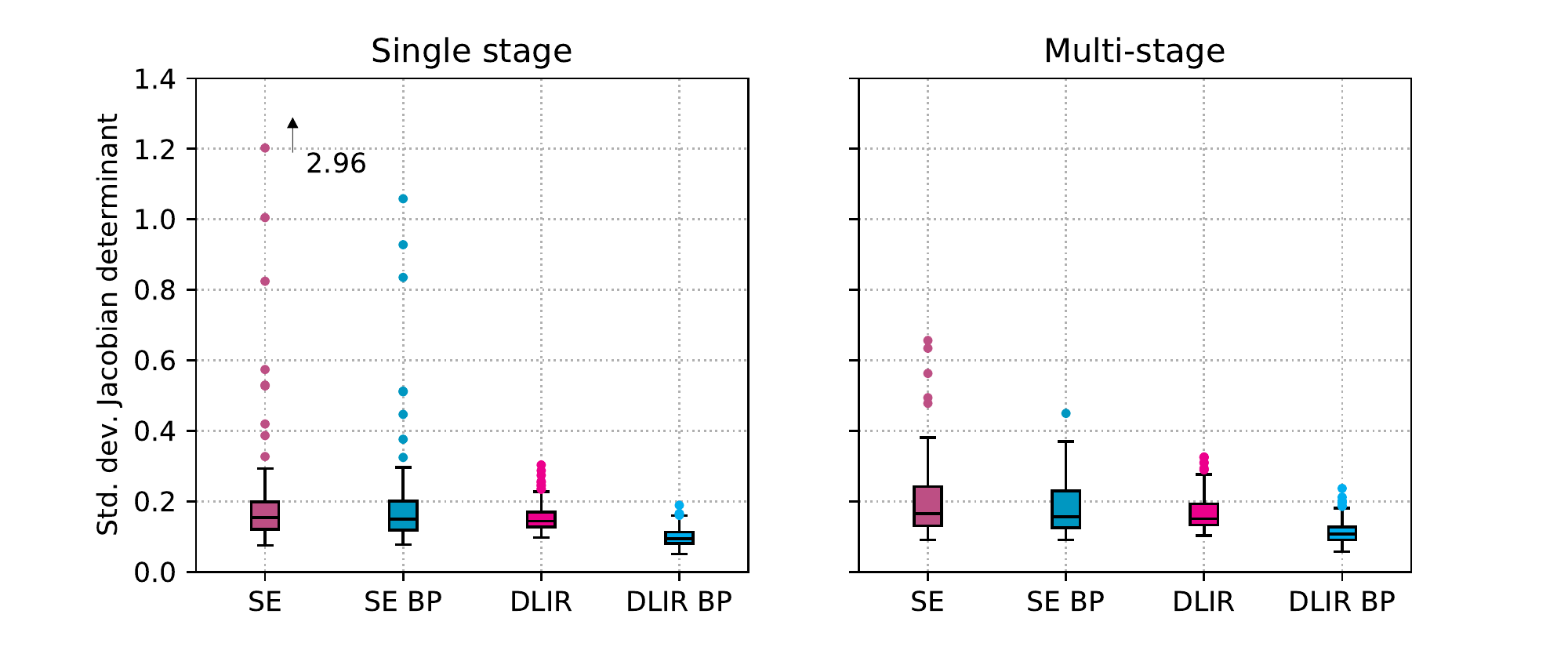}}\\
	\caption{Boxplots showing (a) the volume of singularities and (b) the standard deviation of the Jacobian determinants to evaluate the topology of the DVFs obtained from registration experiments between end-diastole and end-systole cardiac cine MRI. Conventional registration experiments were performed using SimpleElastix (SE) and compared with DLIR registration. Both SE and DLIR experiments were conducted with and without the bending penalty (BP). The necessity of using a mask for conventional registration is illustrated by the results in shown in the single stage experiments. For visualization purposes large outliers are indicated with an arrow with their values annotated.}
	\label{fig:scd_folding_boxplots}
\end{figure*}

\begin{figure*}
\centering
		\subfloat[Dice coefficient]{\label{fig:scd_dice}\includegraphics[width = .9\textwidth]{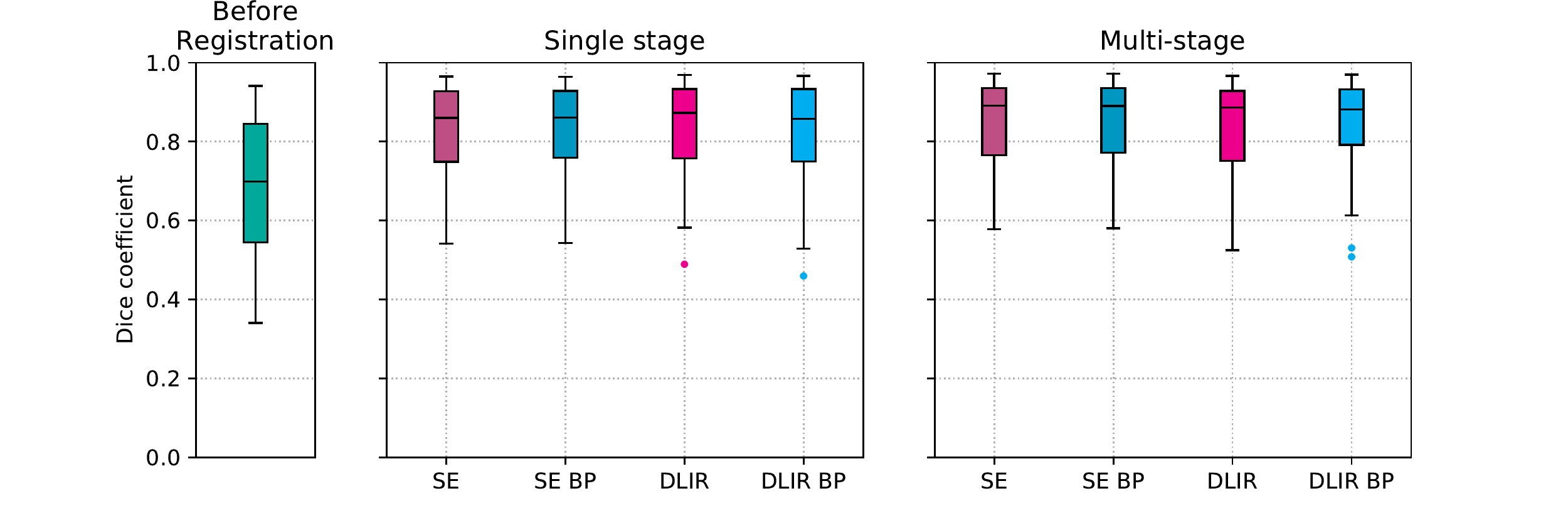}}\\
		\subfloat[Hausdorff distance]{\label{fig:scd_hd}\includegraphics[width = .9\textwidth]{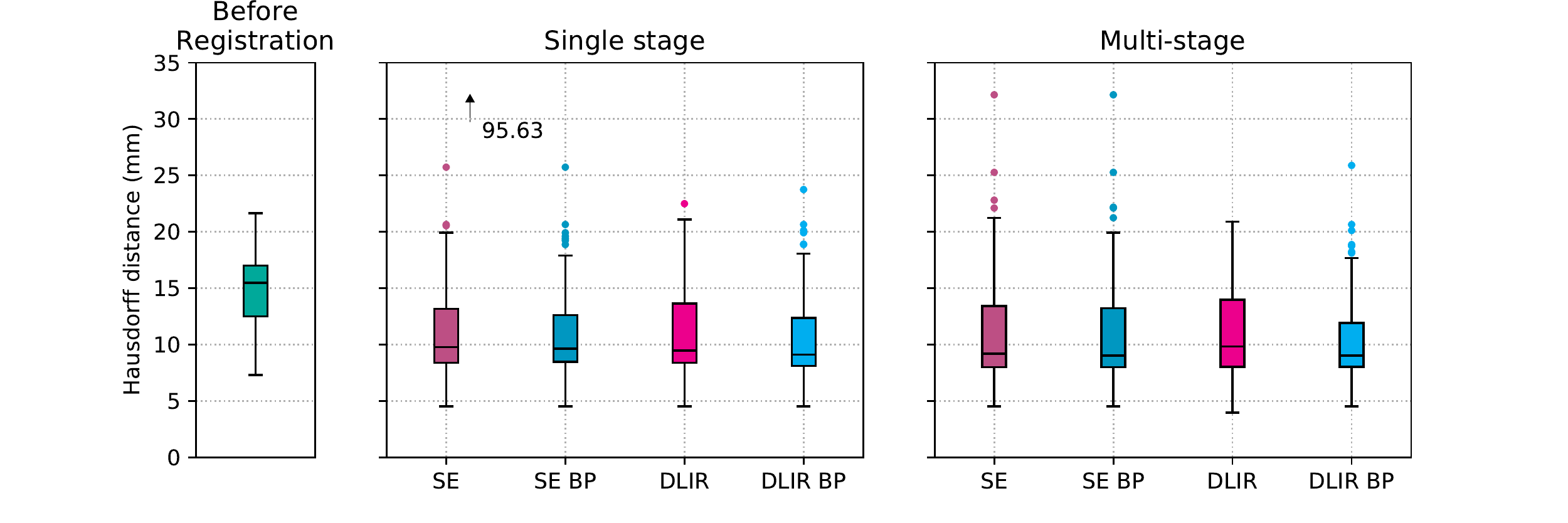}}\\
		\subfloat[Average symmetric surface distance]{\label{fig:scd_asd}\includegraphics[width =  .9\textwidth]{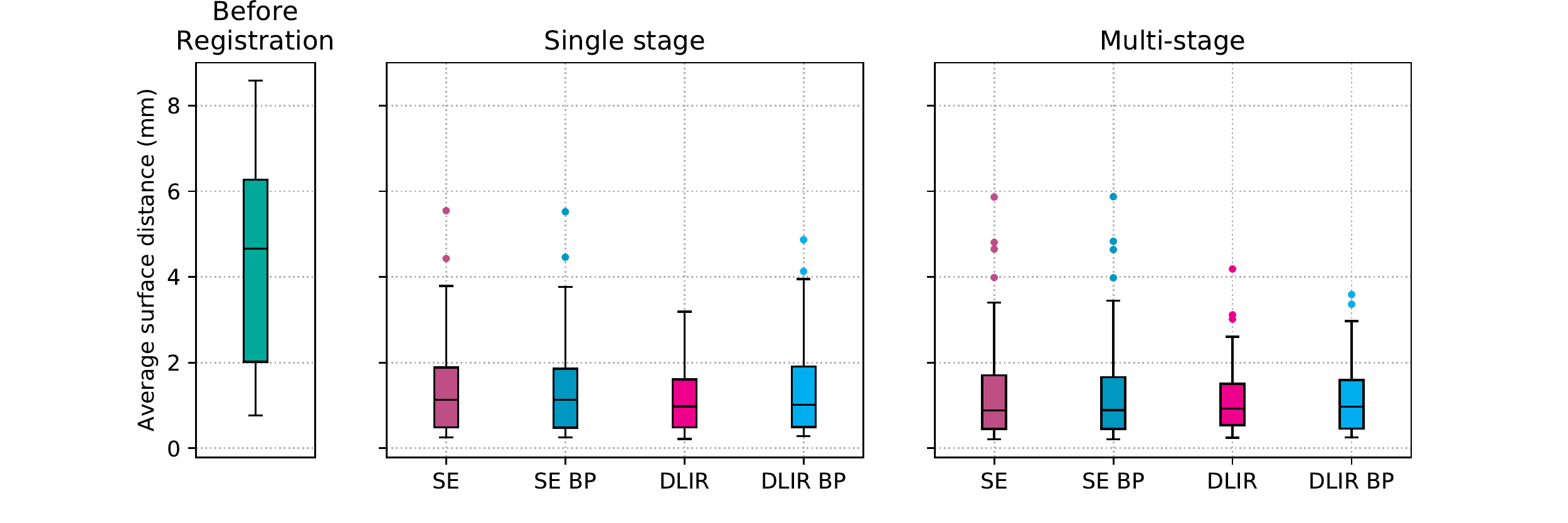}}\\
	\caption{
	Label propagation results of manual left ventricle lumen annotations of intra-patient cardiac cine MRI registration. Boxplots of (a) Dice, (b) Hausdorff distance, and (c) average surface distance are shown for conventional image registration with SimpleElastix (SE) and the DLIR framework.
    Single stage and multi-stage registration experiments were performed for conventional registration and DLIR with and without the bending penalty (BP). The large outlier in (b) was indicated with an arrow to improve visualization.}
	\label{fig:scd_segmentation_boxplots}
\end{figure*}

Figure~\ref{fig:scd_example} shows single stage image registration results of registering images at ES to ED. The obtained Jacobians show that the bending penalty mitigates image folding of the DLIR framework. Furthermore, quantitative analysis, as shown in Figure~\ref{fig:scd_folding_boxplots}, reveals that the DLIR framework is not affected by image folding as much as conventional image registration. Nevertheless, even though nearly absent in the DLIR framework, image folding is further reduced by adding a bending penalty. On the other hand, multi-stage registration seems to have no effect on image folding in the DLIR framework, while having a large effect on folding outliers in conventional image registration. However, the label propagation results, shown in Figure~\ref{fig:scd_segmentation_boxplots}, show that the DLIR framework also benefits from multi-stage image registration. It improved label overlap as indicated by the increased Dice and decreased ASD. The Hausdorff distance appears to be similar across experiments.

\newcommand{\cardiacthumbwidth}{0.225\textwidth}
\begin{figure*}
	\centering
	\setlength\tabcolsep{1.5pt}
	\begin{tabular}{cc}

		\small Fixed Image &
		\small Moving Image\\ 
		\includegraphics[width = \cardiacthumbwidth]{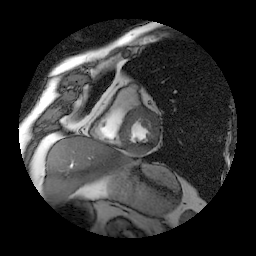}&
    
    	\includegraphics[width = \cardiacthumbwidth]{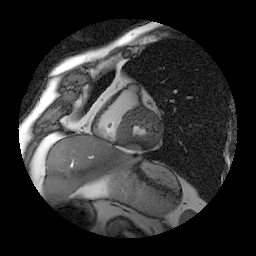}\\

	\end{tabular}
	\centering
	\begin{tabular}{m{\cardiacthumbwidth}m{\cardiacthumbwidth}m{\cardiacthumbwidth}m{\cardiacthumbwidth}}
		\multicolumn{1}{c}{\small SE} & 
		\multicolumn{1}{c}{\small SE + BP} & 
		\multicolumn{1}{c}{\small DLIR}&
		\multicolumn{1}{c}{\small DLIR + BP}
        \\

        \includegraphics[width = \cardiacthumbwidth]{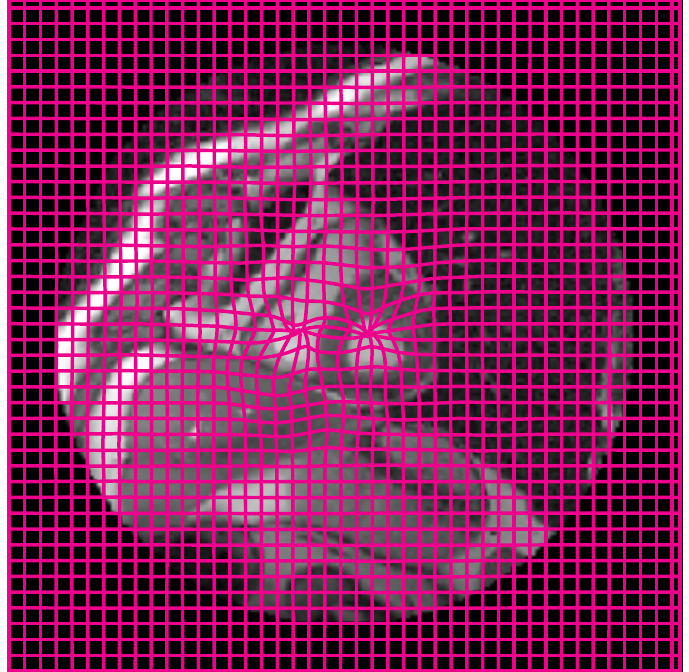}&
        \includegraphics[width = \cardiacthumbwidth]{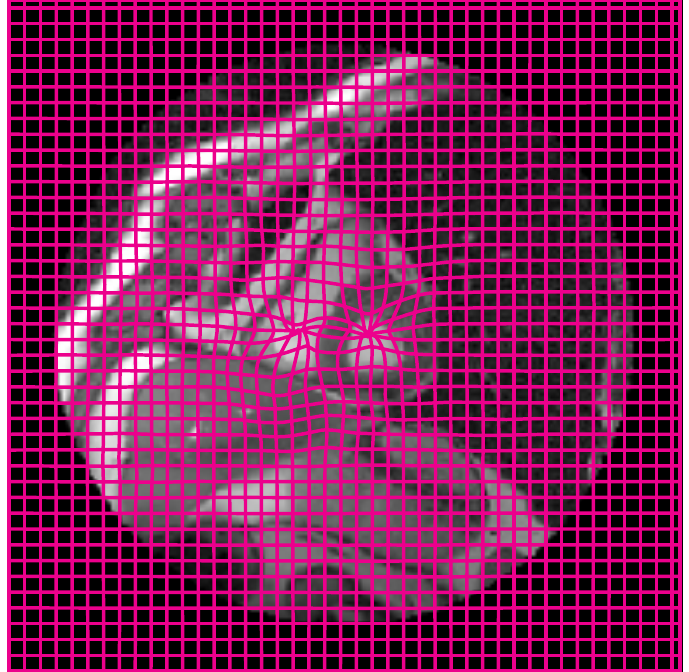}&
        \includegraphics[width = \cardiacthumbwidth]{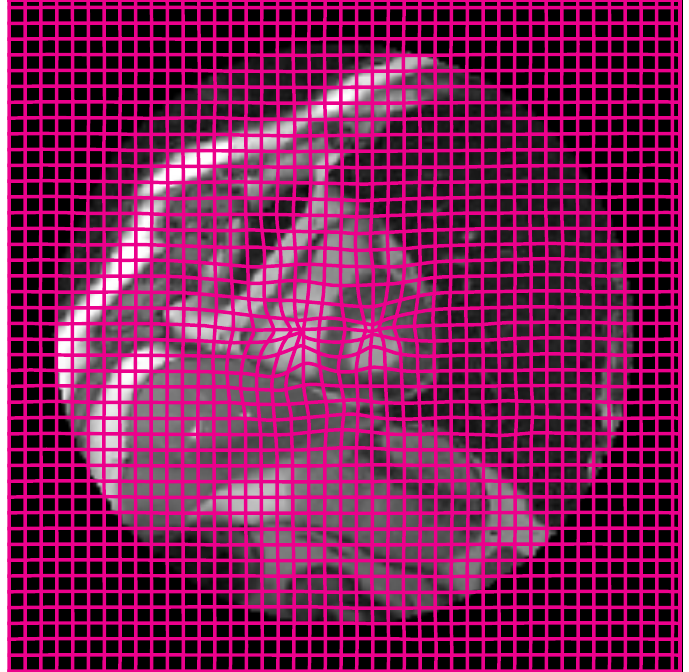}&
        \includegraphics[width = \cardiacthumbwidth]{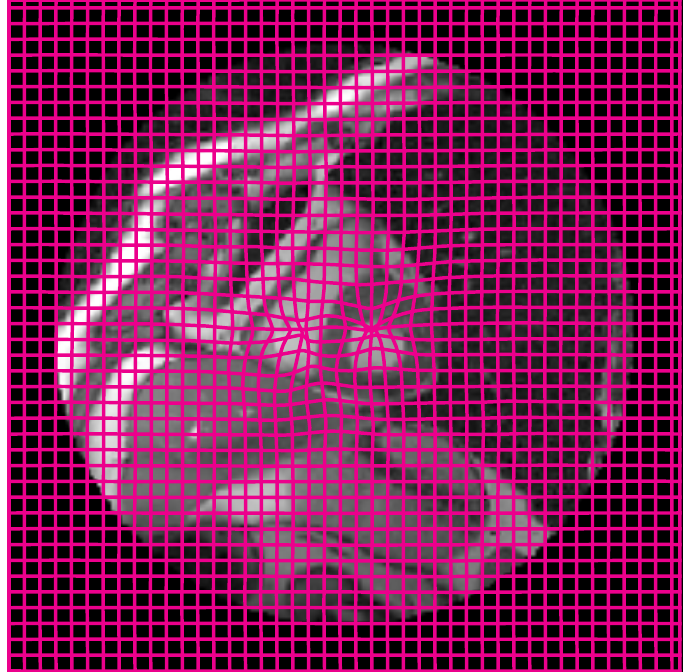}
        \\

        \includegraphics[width = \cardiacthumbwidth]{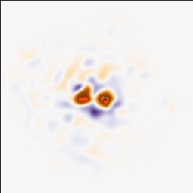}&
        \includegraphics[width = \cardiacthumbwidth]{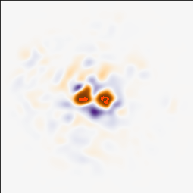}&
        \includegraphics[width = \cardiacthumbwidth]{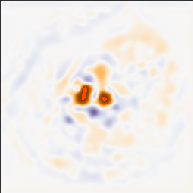}&
        \includegraphics[width = \cardiacthumbwidth]{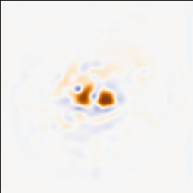}
        \\
        \multicolumn{4}{c}{\includegraphics[width=.4\textwidth]{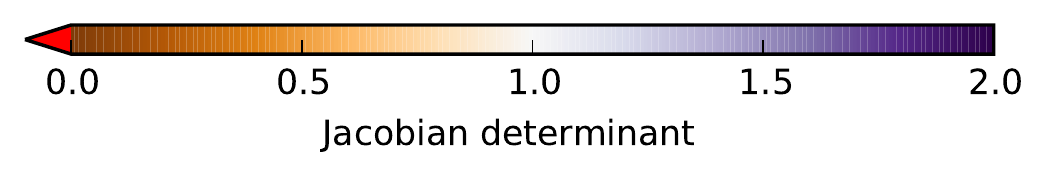}}
	\end{tabular}
	\caption{Top row: Cardiac cine MRI of a patient with left ventricular hypertrophy. Center axial slices are taken from the end-diastolic time point (Fixed) and end-systolic (Moving) timepoints. For visualization purposes fixed and moving images are cropped to the heart. Middle row: Registration results with superpositioned deformation grids. Bottom row: Colormap of the Jacobian with singularities (folding) indicated in bright red. From left to right results are shown for SimpleElastix (SE) and DLIR, with and without the bending penalty (BP).}
	\label{fig:scd_example}
\end{figure*}

\begin{figure*}
\centering
		\subfloat[Single stage registration]{\includegraphics[width = .4\textwidth]{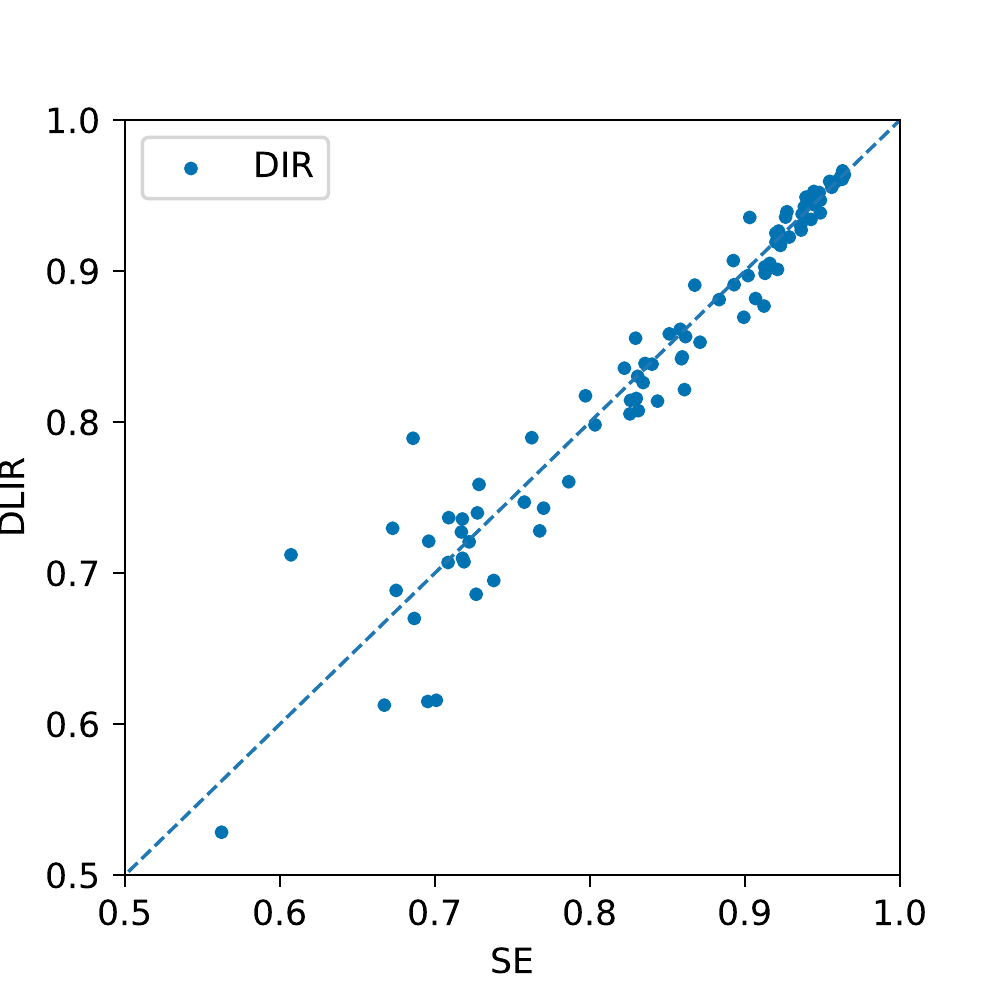}}
		\subfloat[Multi-stage registration]{\includegraphics[width = .4\textwidth]{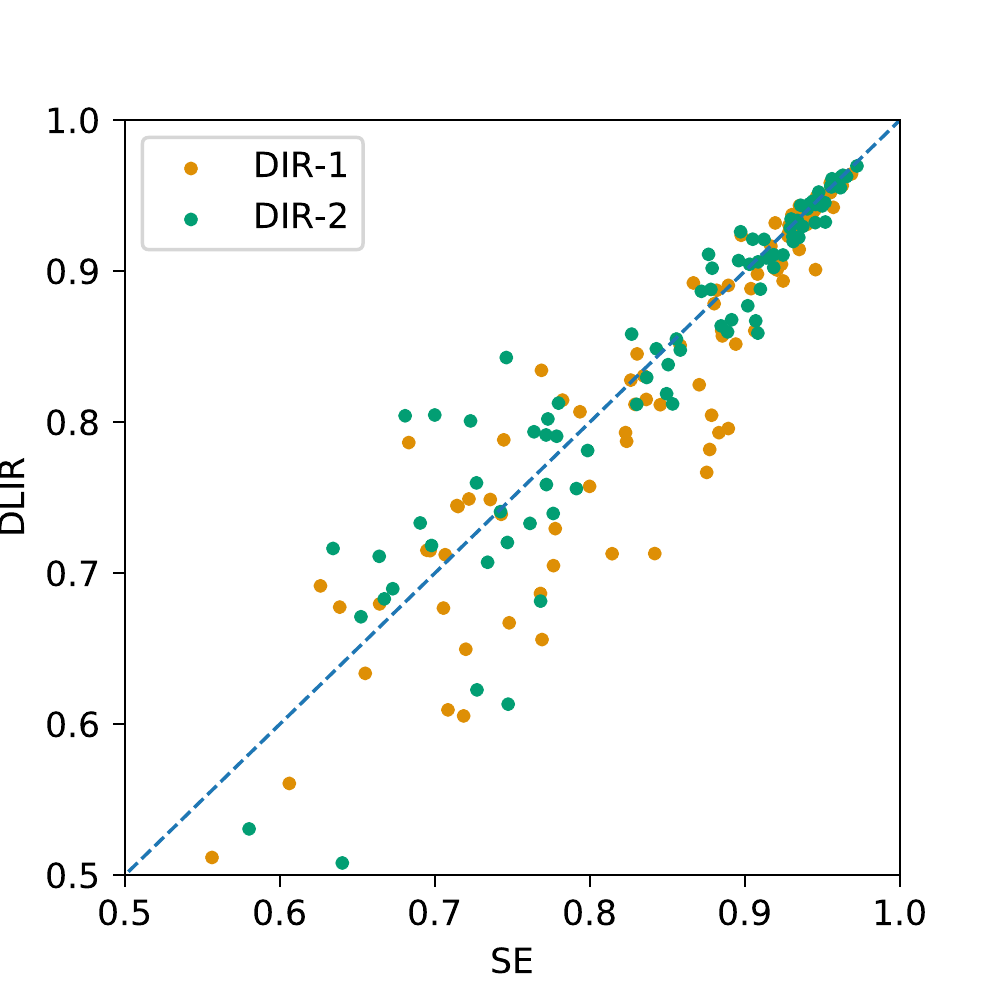}}
	\caption{Scatter plots showing a comparison of Dice scores obtained with the DLIR framework and conventional inter-patient cardiac cine MRI registration. The plots show a correlation, but the dispersion of the points indicate that the registration tasks are not equally difficult for the DLIR framework and conventional registration framework.}
	\label{fig:scatterplots:scd}
\end{figure*}

Figure~\ref{fig:scatterplots:scd} provides additional insight into registration performance of DLIR vs. conventional image registration. The spread shows that there is no correlation between frameworks with respect to registration results of image pairs; some image pairs were well aligned with DLIR and poorly with the conventional approach, and vice versa.

Table~\ref{tab:scd_results} provides an overview of all results. Statistical analyses with the Wilcoxon signed rank test indicated that the multi-stage DLIR with bending penalty had significantly less folding and lower standard deviation of the Jacobians ($p\ll0.0001$) compared to other methods. Dice and ASD were as high as conventional image registration and significantly better compared to single stage experiments. Interestingly, the multi-stage DLIR is approximately 350 times faster than the single stage conventional image registration experiments, and takes only 39\,ms for multi-stage image registration, including intermediate and final image resampling.

\begin{table*}
\caption{Table listing the results of cardiac cine MRI registration experiments. Single stage and multi-stage conventional and DLIR registration are compared with and without bending penalties (BP). Given that the results are not following a normal distribution, median $\pm$ interquartile ranges are provided. Execution times are provided as mean (standard deviation). Note that the bending penalty is only applied to the DLIR framework during training, thus during application it does not limit execution time.}
\label{tab:scd_results}
\resizebox{\textwidth}{!}{%
\begin{tabular}{ll|ccccccc}
                              &            & Dice        & HD           & ASD         & Fraction folding & Std. dev.  Jacobian & CPU time (s) & GPU time (s)                      \\
\hline
\multicolumn{2}{l|}{Before registration}\Tstrut    & $ 0.70\pm0.30 $ & $ 15.46\pm4.50 $ & $ 4.66\pm4.26 $ &--       &  --     &  --   &         --                   \\
\hline\Tstrut

    \multirow{4}{*}{Single stage}   & SE    & $ 0.86\pm0.18 $ & $ 9.76\pm4.78 $  & $ 1.14\pm1.40 $ & $ 0.08\pm0.16 $  & $ 0.15\pm0.08 $     &  $ 13.49(3.27) $     &  --                \\
                              & SE + BP & $ 0.86\pm0.17 $ & $ 9.64\pm4.15 $  & $ 1.13\pm1.38 $ & $ 0.07\pm0.15 $   & $ 0.15\pm0.08 $     &        $ 14.89(3.07) $     & --                  \\
\cline{3-9}
                              & DLIR       & $ 0.87\pm0.18 $ & $ 9.47\pm5.26 $  & $ 0.98\pm1.12 $ & $ 0.03\pm0.06 $  & $ 0.14\pm0.04 $     & \multirow{2}{*}{$\phantom{1}1.71(0.45)$} & \multirow{2}{*}{$0.03\pm0.01 $}\\ 
                              & DLIR + BP  & $ 0.86\pm0.18 $ & $ 9.10\pm4.26 $  & $ 1.01\pm1.42 $ & $ 0.00\pm0.01 $  & $ 0.09\pm0.03 $     &                                \\
\hline\Tstrut
\multirow{4}{*}{Multi-stage}  & SE    & $ 0.89\pm0.17 $ & $ 9.18\pm5.42 $  & $ 0.88\pm1.25 $ & $ 0.08\pm0.17 $  & $ 0.17\pm0.11 $     &  $ 15.51(3.67) $     & --               \\
                              & SE + BP & $ 0.89\pm0.16 $ & $ 9.01\pm5.23 $  & $ 0.89\pm1.21 $ & $ 0.05\pm0.16 $  & $ 0.16\pm0.11 $     &  $ 20.06(3.68) $     & --                   \\                          
                              
\cline{3-9}
                              & DLIR       & $ 0.89\pm0.18 $ & $ 9.84\pm5.93 $  & $ 0.93\pm0.97 $ & $ 0.05\pm0.08 $  & $ 0.15\pm0.06 $     & \multirow{2}{*}{$\phantom{1}2.35(0.60)$} &\multirow{2}{*}{$0.04(0.01) $} \\
                              & DLIR + BP  & $ 0.88\pm0.14 $ & $ 9.01\pm3.89 $  & $ 0.97\pm1.14 $ & $ 0.002\pm0.03 $  & $ 0.11\pm0.04 $     &                               
\end{tabular}
}
\end{table*}

\FloatBarrier
\section{Inter-Patient registration of Low-Dose Chest CT}
Inter-patient registration was performed with chest CT scans of different subjects from the NLST. In this set large variations in the field of view were present, which were caused by differences in scanning protocol and by the different CT-scanners that were used. Because of these variations, and the variations in anatomy among subjects, affine registration was necessary for initial alignment. 
Therefore, multi-stage image registration was performed with sequential affine and deformable image registration stages. 
The test-scans provided 90 registrations for evaluation. Manual delineation of the aorta and 10 landmarks were used to assess registration performance.

\subsection{ConvNet Design and Training}
Inter-patient chest-CT registration requires initial alignment of patient scans. Thus, we implemented a multi-stage ConvNet consisting of an affine registration stage, followed by coarse-to-fine deformable image registration. The full Hounsfield Unit range of CT numbers (-1000 to 3095) was used to rescale input image intensities from 0~to~1.
Memory limitations imposed by hardware and software limited the deformable image registration to three stages and a final image resolution of 2\,mm. In-plane slice sizes ranged from 115$\times$115 to 250$\times$250 voxels and the number of slices ranged from 109 to 210. All ConvNets were designed with 32 kernels per convolution layer, but downsampling was performed with strided convolutions with $2\times2\times2$ downsampling and $4\times4\times4$ kernels instead of the favorable average pooling \citep{DeVos2017} to further limit memory consumption. The affine registration ConvNet was designed as shown in Figure~\ref{fig:affine}, but with three downsampling layers in the pipelines. The separate pipelines in the affine registration ConvNet allowed analysis of a fixed and a moving image having different dimensions. The affine ConvNet registers the moving image to the fixed image space. As a result, the fixed and moving pairs can be concatenated and used for subsequent deformable image registration ConvNets, which were designed as specified in Figure~\ref{fig:freeform}.

\begin{figure*}
	\centering
	\includegraphics[width=.75\textwidth]{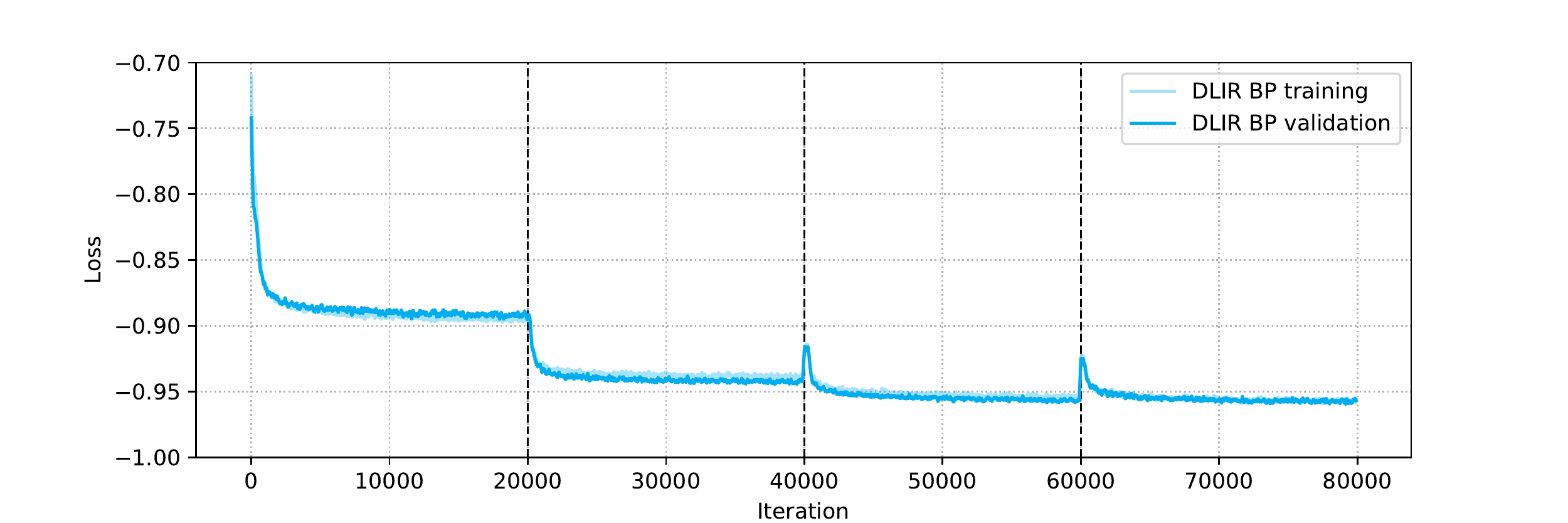}
	\caption{Learning curves during sequential training of the four registration stages of a ConvNet for inter-patient registration of chest~CT.}
	\label{fig:nlst_learning_curves}
\end{figure*}

The multi-stage ConvNet was trained in mini-batches consisting of randomly selected image pairs. Given that the training set consisted of 2,000 scans, almost four million possible permutations of image pairs were available for training. Not all permutations were seen during training, but on average each scan was analyzed 674 times. Additionally, random augmentations were performed by randomly cropping 32\,mm in any direction.
The multi-stage ConvNet was trained in 18 hours using the settings listed in Table~\ref{tab:nlst_settings}. The loss curves shown in Figure~\ref{fig:nlst_learning_curves} show no signs of overfitting. The third and fourth stages analyze higher resolution images and output finer B-spline grids. As a consequence the dissimilarity increases and the finer deformations increase the bending penalty, resulting in higher starting losses, compared to previous registration stages.

\begin{table*}[]
	\centering
	\caption{Experimental settings of the DLIR framework for training a multi-stage ConvNet for inter-patient registration of chest CT. The ConvNet consists of an affine image registration (AIR) stage, and three deformable image registration (DIR) stages. Image resolution, grid spacing, and average number of grid points are given in x$\times$y$\times$z order.}
	\label{tab:nlst_settings}
	\begin{tabular}{lccccc}
		Stage & AIR    & DIR-1    & DIR-2    & DIR-3    \\
		\hline\Tstrut
		Input image resolution (mm)                    & $8\times8\times8$      & $8\times8\times8$      & $4\times4\times4$     & $2\times2\times2$     \\

		Grid spacing (mm)                    & --     & $64\times64\times64$     & $32\times32\times32$     & $16\times16\times16$     \\
		Avg. grid points & --  & $5\times5\times5$ & $11\times11\times10$ & $21\times21\times20$\\
		Mini-batch size (pairs)              & $16$     & $8$      & $4$      & $2$   \\

	\end{tabular} %
\end{table*}

\subsection{Results}
Ten images with manually segmented aortas resulted in 90 permutations of fixed and moving image pairs that were used for evaluation. Figure~\ref{fig:nlst_example} shows that the affine stage correctly aligns two images from the evaluation set. The coarse-to-fine deformable stages gradually improves upon this alignment. However, final DVFs obtained in these experiment show some folding, as is visualized in the examples of Figure~\ref{fig:nlst_example2}.

\newcommand{\chestthumbwidth}{0.15\textwidth}
\begin{figure*}
	\centering
	\setlength\tabcolsep{1.5pt}
	\begin{tabular}{m{\chestthumbwidth}m{\chestthumbwidth}m{\chestthumbwidth}m{\chestthumbwidth}m{\chestthumbwidth}m{\chestthumbwidth}}
		\multicolumn{1}{c}{Moving} & 
		\multicolumn{1}{c}{AIR} & 
		\multicolumn{1}{c}{DIR-1} & 
		\multicolumn{1}{c}{DIR-2} & 
		\multicolumn{1}{c}{DIR-3} & 

		\multicolumn{1}{c}{Fixed}\\
		\includegraphics[width = \chestthumbwidth]{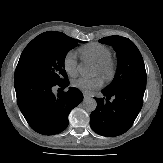}&
		\includegraphics[width = \chestthumbwidth]{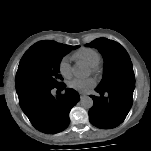}&
		\includegraphics[width = \chestthumbwidth]{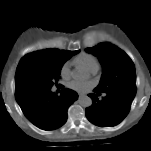}&
		\includegraphics[width = \chestthumbwidth]{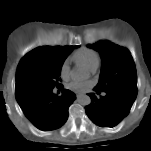}&
		\includegraphics[width = \chestthumbwidth]{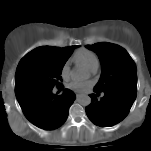}&

		\includegraphics[width = \chestthumbwidth]{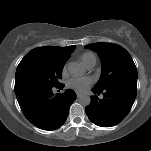}\\
		\includegraphics[width = \chestthumbwidth]{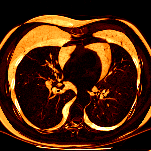}&
		\includegraphics[width = \chestthumbwidth]{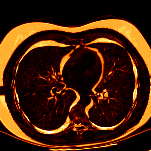}&
		\includegraphics[width = \chestthumbwidth]{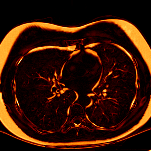}&
		\includegraphics[width = \chestthumbwidth]{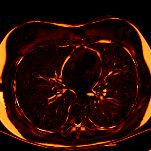}&
		\includegraphics[width = \chestthumbwidth]{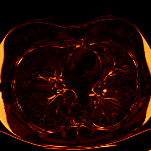}&

		\\
		\includegraphics[width = \chestthumbwidth, trim={0 0.25cm 0 0.25cm}, clip]{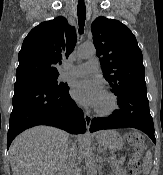}&
		\includegraphics[width = \chestthumbwidth]{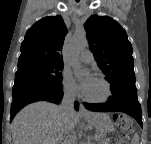}&
		\includegraphics[width = \chestthumbwidth]{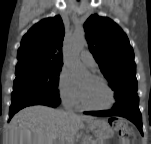}&
		\includegraphics[width = \chestthumbwidth]{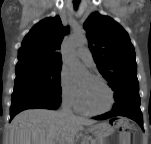}&
		\includegraphics[width = \chestthumbwidth]{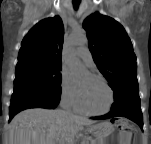}&

		\includegraphics[width = \chestthumbwidth]{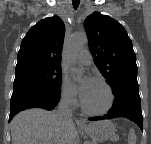}\\
		\includegraphics[width = \chestthumbwidth]{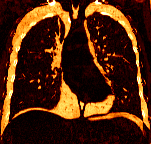}&
		\includegraphics[width = \chestthumbwidth]{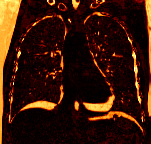}&
		\includegraphics[width = \chestthumbwidth]{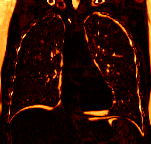}&
		\includegraphics[width = \chestthumbwidth]{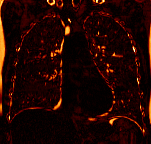}&
		\includegraphics[width = \chestthumbwidth]{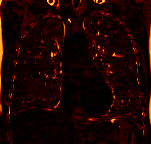}&

		\\
		\includegraphics[width = \chestthumbwidth, trim={0 0.25cm 0 0.25cm}, clip]{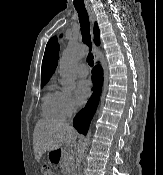}&
		\includegraphics[width = \chestthumbwidth]{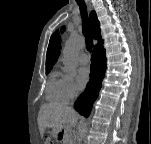}&
		\includegraphics[width = \chestthumbwidth]{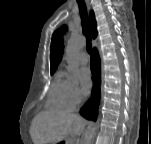}&
		\includegraphics[width = \chestthumbwidth]{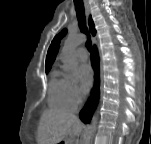}&
		\includegraphics[width = \chestthumbwidth]{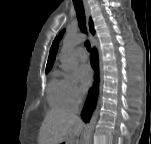}&

		\includegraphics[width = \chestthumbwidth]{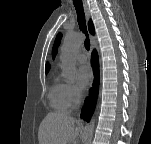}\\
		\includegraphics[width = \chestthumbwidth]{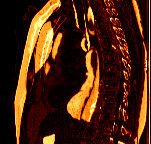}&

		\includegraphics[width = \chestthumbwidth]{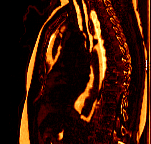}&
		\includegraphics[width = \chestthumbwidth]{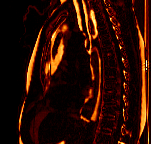}&
		\includegraphics[width = \chestthumbwidth]{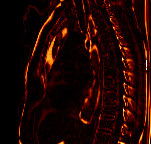}&
		\includegraphics[width = \chestthumbwidth]{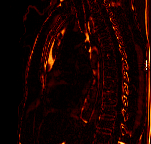}&

		\\
	\end{tabular}
	\caption{Example results of inter-patient registration of two chest CTs from the NLST test set. The moving image is shown on the left and the target fixed image is shown on the right. Intermediate registration results for each stage are shown in between. The rows show center slices of resp. axial, coronal, and sagittal planes, with in between corresponding heatmaps of the absolute difference with respect to the fixed image. This qualitatively shows increasing alignment at each registration stage. The full scale of Hounsfield units cannot be visualized. Window and level is set to visualize the aorta. As a consequence, complexity of the lungs is not visible in this example.}
	\label{fig:nlst_example}
\end{figure*}

\newcommand{\chestthumbwidthtwo}{0.225\textwidth}
\begin{figure*}
	\centering
	\setlength\tabcolsep{0.5pt}	
	\begin{tabular}{ccccc}
    Fixed & Warped & Moving & Jacobian  \\
	 
        \includegraphics[width = \chestthumbwidthtwo]{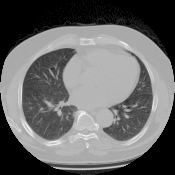}&
        \includegraphics[width = \chestthumbwidthtwo]{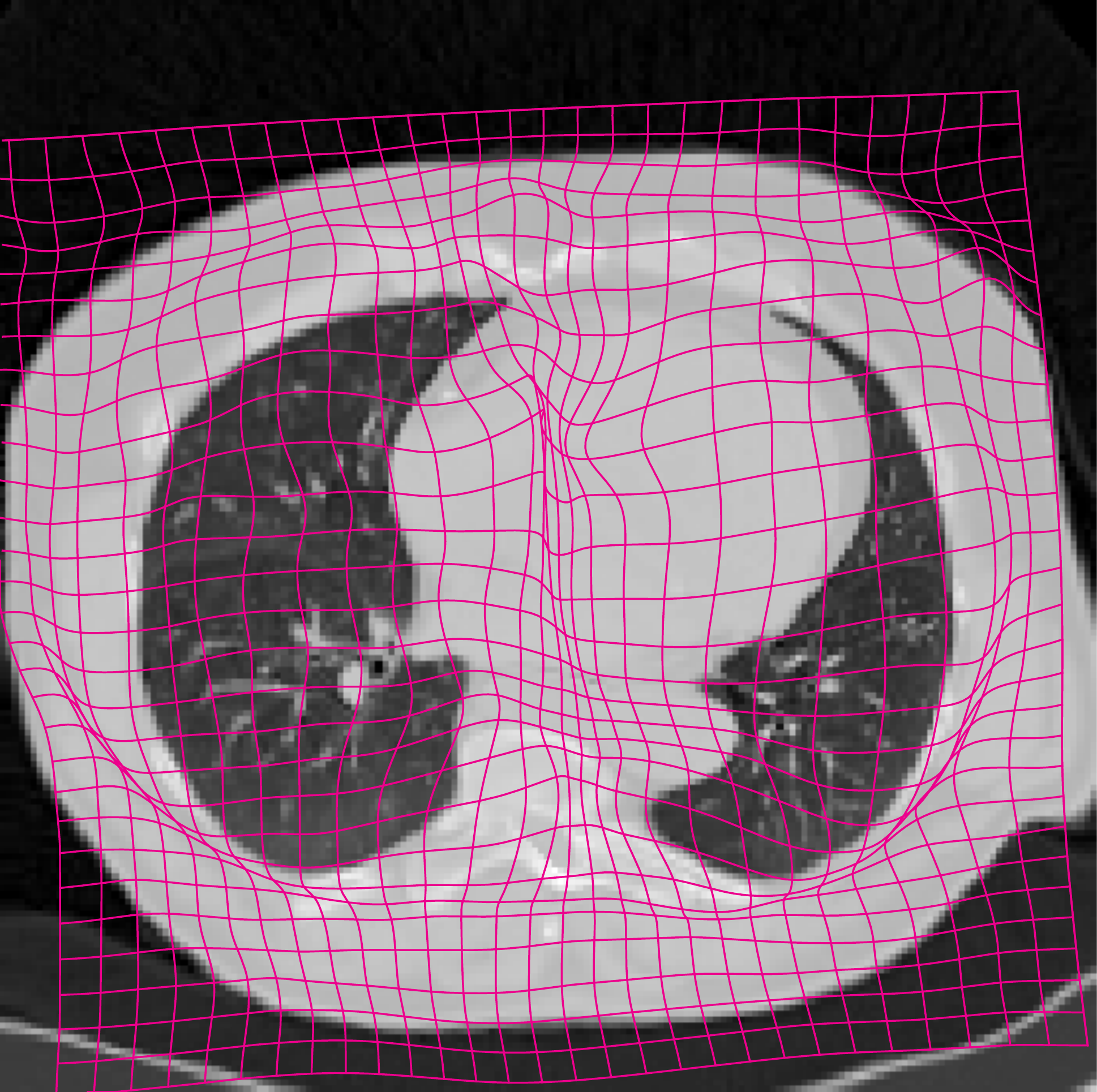}&
        \includegraphics[width = \chestthumbwidthtwo]{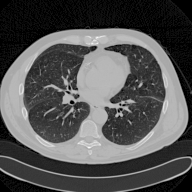}&
        \includegraphics[width = \chestthumbwidthtwo]{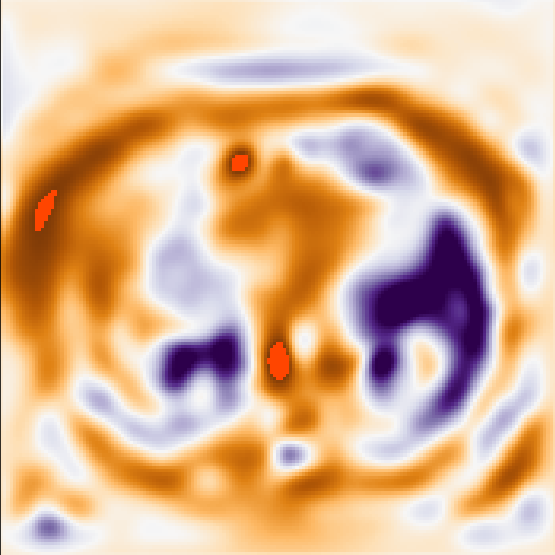}&
        \multirow{4}{*}{\includegraphics[height=.5\textwidth]{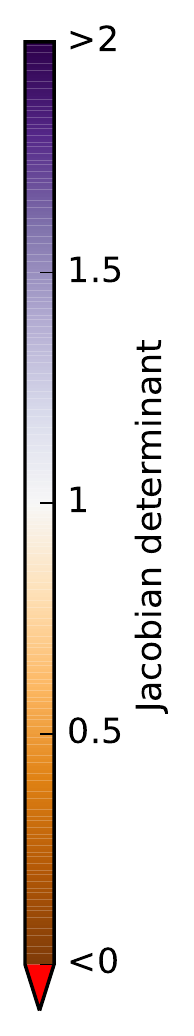}}\\
        
        \includegraphics[width = \chestthumbwidthtwo]{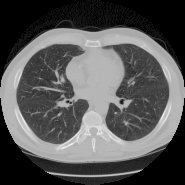}&
        \includegraphics[width = \chestthumbwidthtwo]{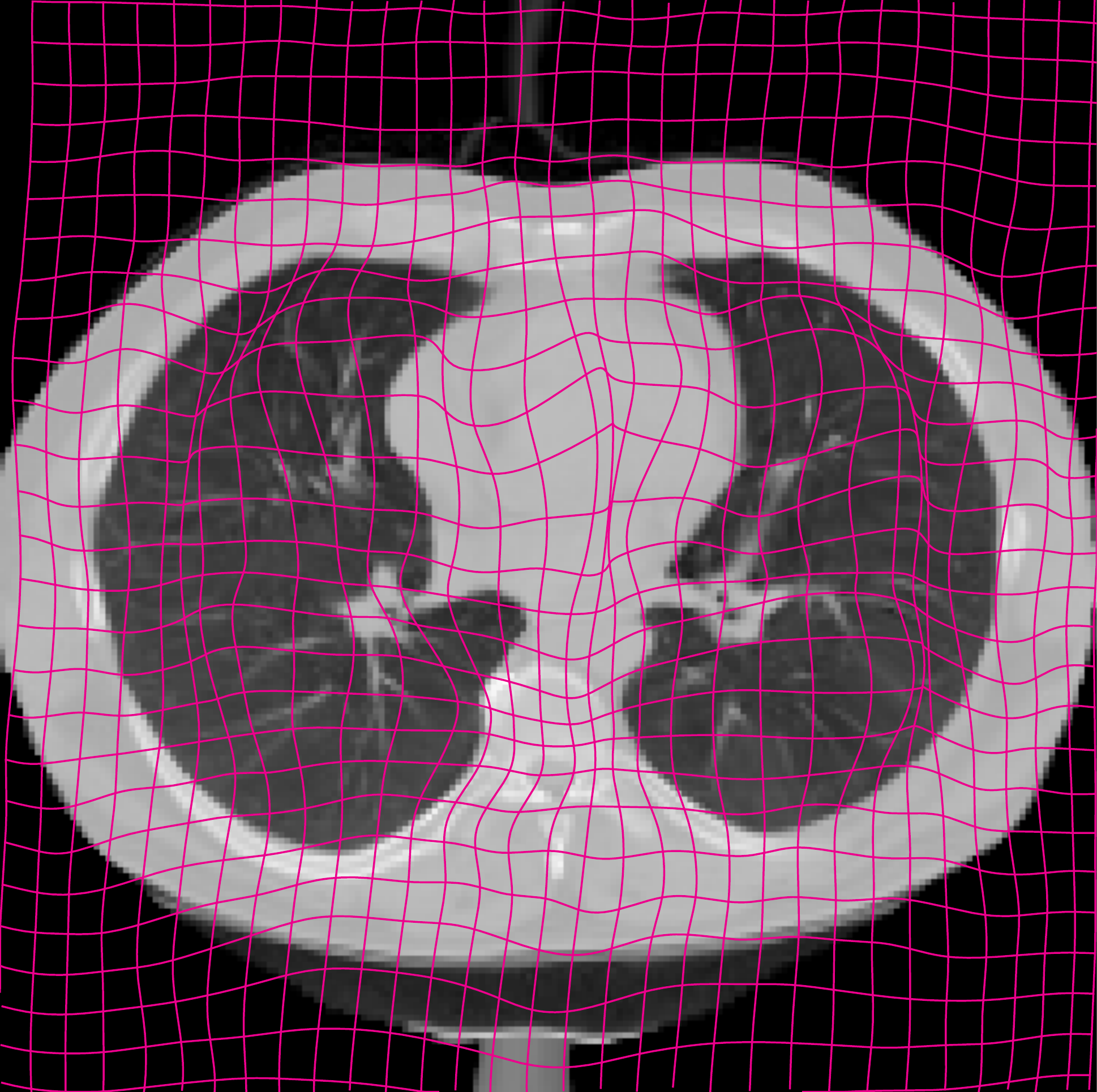}&
        \includegraphics[width = \chestthumbwidthtwo]{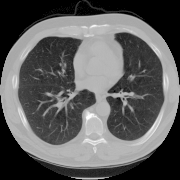}&
        \includegraphics[width = \chestthumbwidthtwo]{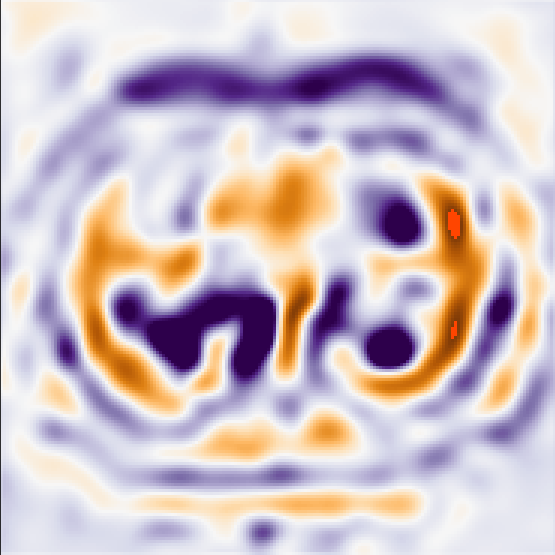}\\
        
        \includegraphics[width = \chestthumbwidthtwo]{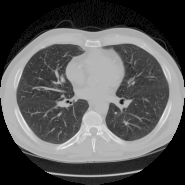}&
        \includegraphics[width = \chestthumbwidthtwo]{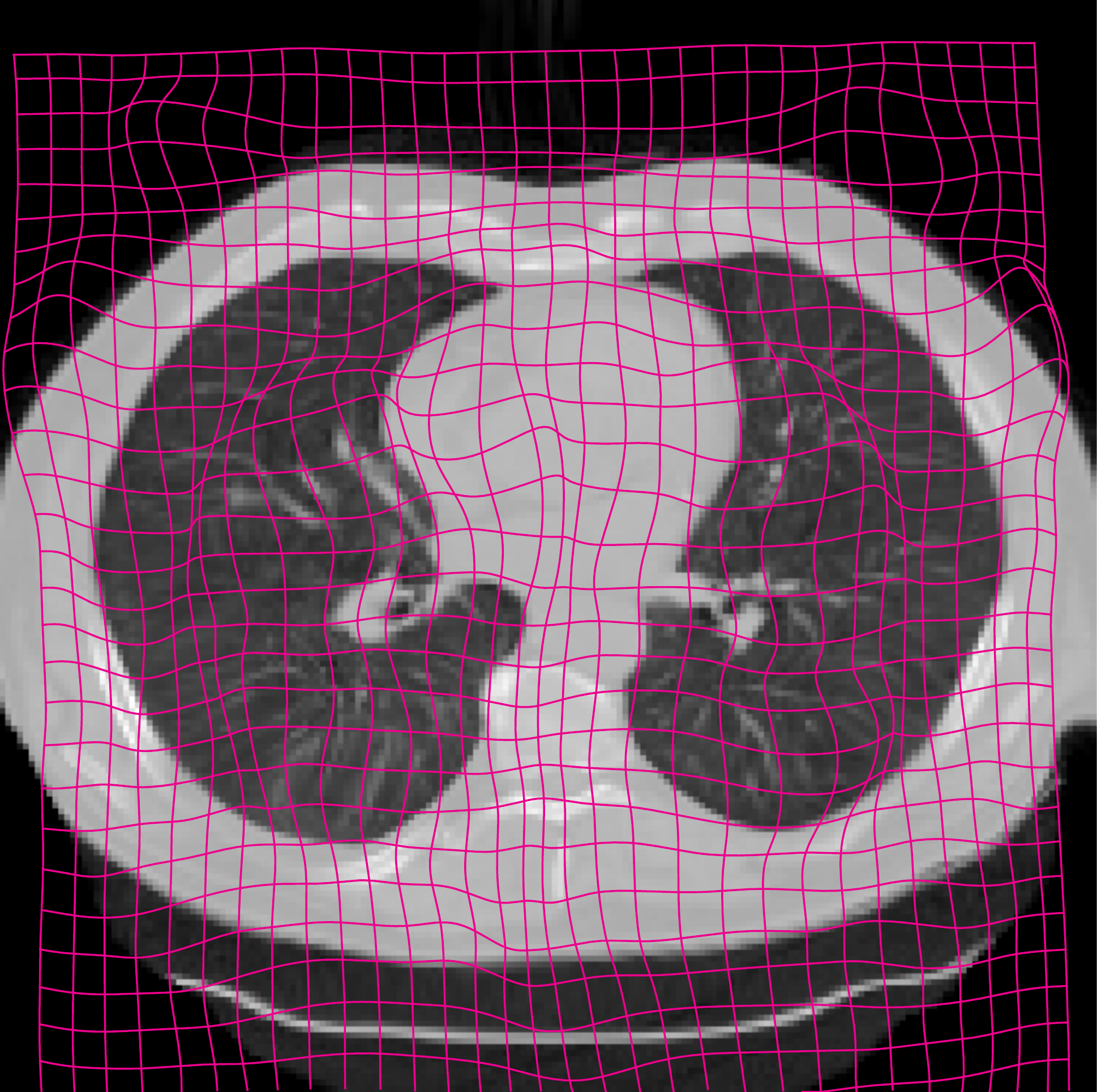}&
        \includegraphics[width = \chestthumbwidthtwo]{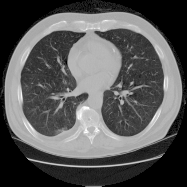}&
        \includegraphics[width = \chestthumbwidthtwo]{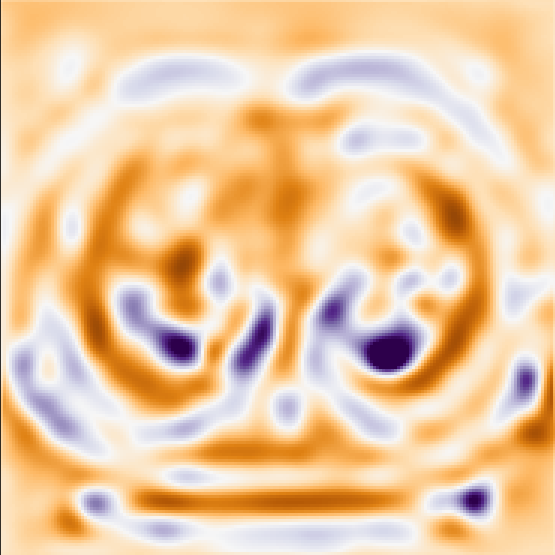}\\
        
        \includegraphics[width = \chestthumbwidthtwo]{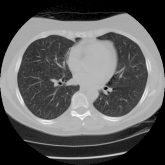}&
        \includegraphics[width = \chestthumbwidthtwo]{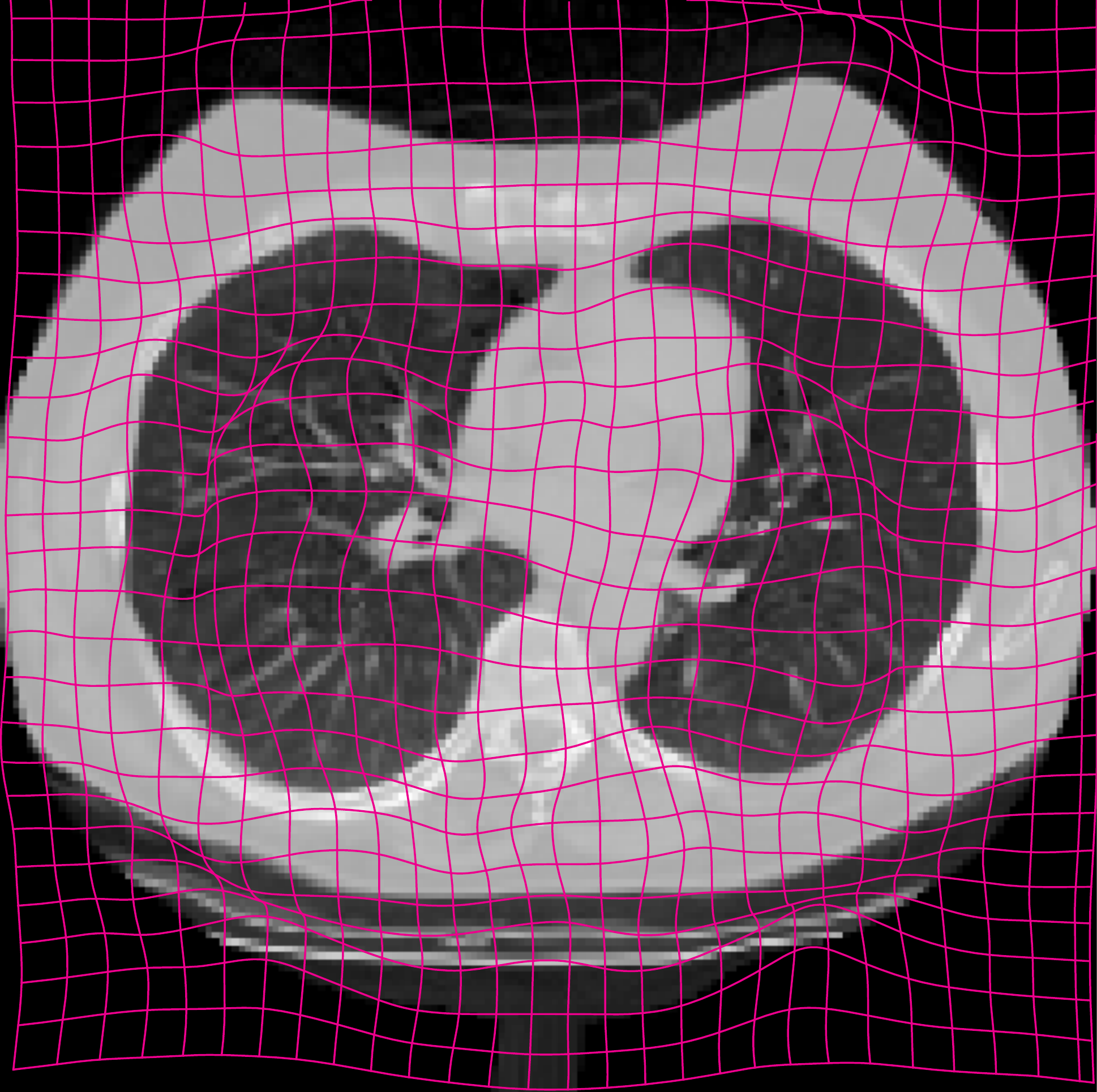}&
        \includegraphics[width = \chestthumbwidthtwo]{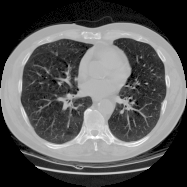}&
        \includegraphics[width = \chestthumbwidthtwo]{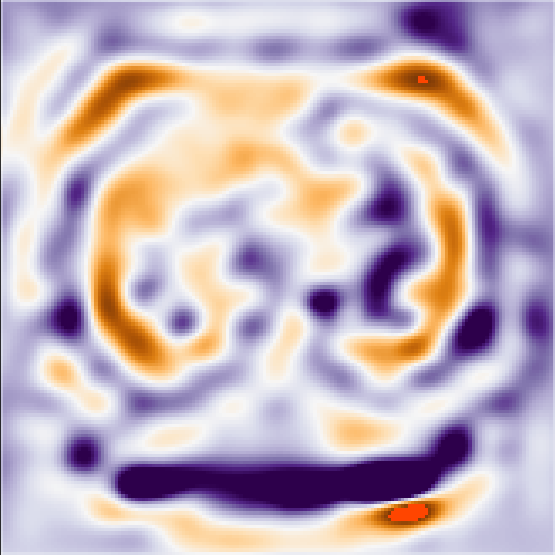}\\
        
	\end{tabular}
	\caption{The rows show four inter patient chest-CT registration results. The columns show fixed images, warped images with a deformation grid, moving images, and a colormap of the Jacobian with singularities (folding) indicated in bright red.}
	\label{fig:nlst_example2}
\end{figure*}

\begin{figure*}
\centering
        \subfloat[Amount of folding]{\label{fig:nlst_frac_folding}\includegraphics[width = .4\textwidth]{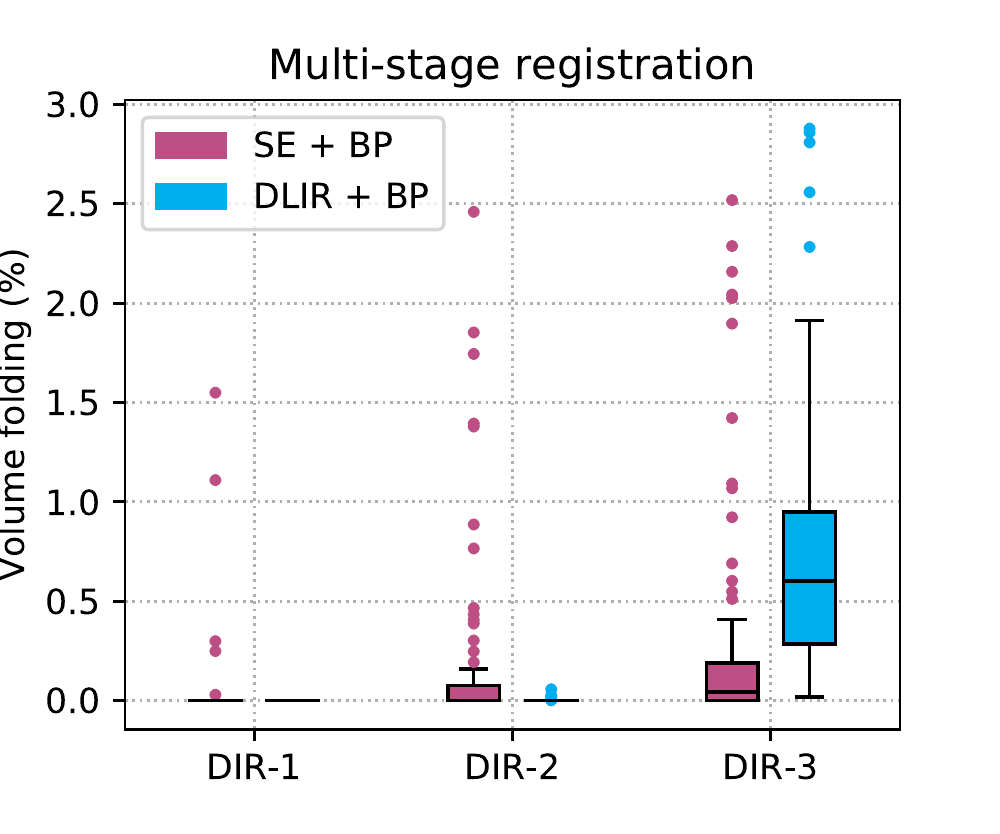}}
		\subfloat[Standard deviation of Jacobian]{\label{fig:nlst_jac_std}\includegraphics[width = .4\textwidth]{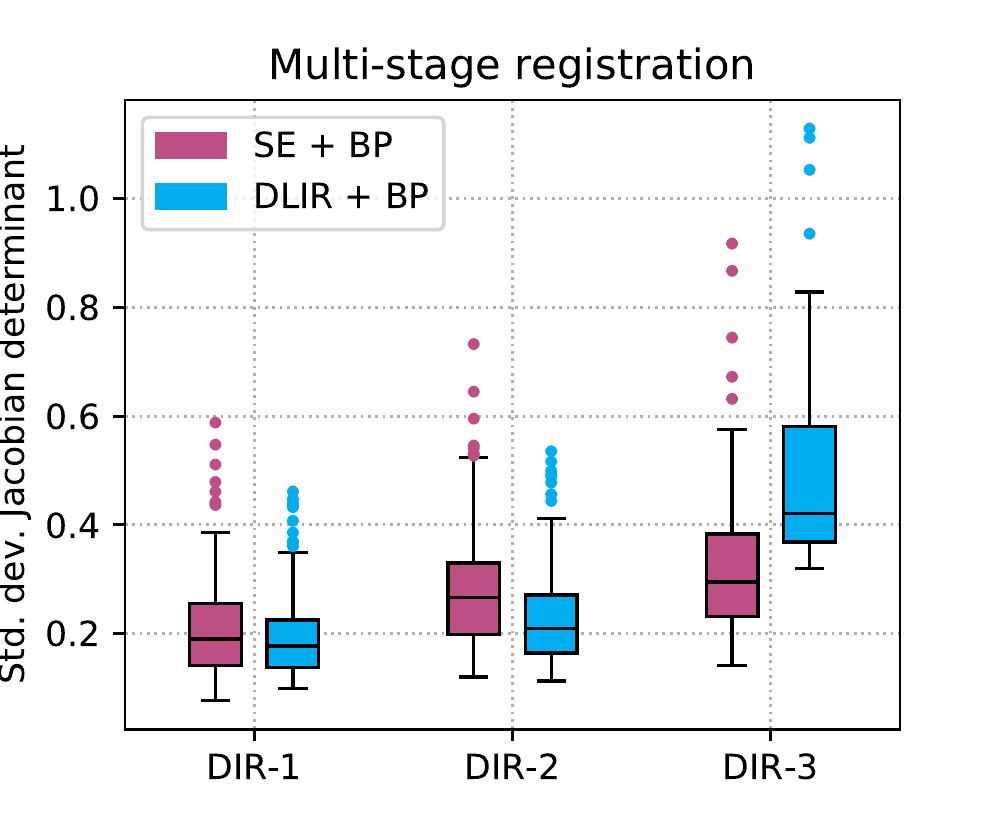}}\\
	\caption{Boxplots showing in (a) the volume fraction of folding and in (b) the standard deviation of the Jacobian determinants of the deformable stages of inter-patient chest CT registration. Conventional registration experiments were performed using SimpleElastix (SE) and compared with DLIR registration both using a bending penalty (BP).}
	\label{fig:nlst_foldings}
\end{figure*}

Quantitative analysis, shown in Figure~\ref{fig:nlst_foldings}, of the deformable registration DVFs reveals that only the final registration stage is hampered by folding. While in conventional image registration folding gradually increases with each deformable registration stage, the DLIR framework shows zero to limited folding in the first two stages and a large increase in the final stage. A similar pattern is seen in the standard deviations of the Jacobians. The Wilcoxon signed-rank test indicated that for each stage the results were significantly different between conventional image registration and the DLIR framework.

\begin{figure*}
\centering
        \subfloat[Dice coefficient]{\includegraphics[width = .5\textwidth]{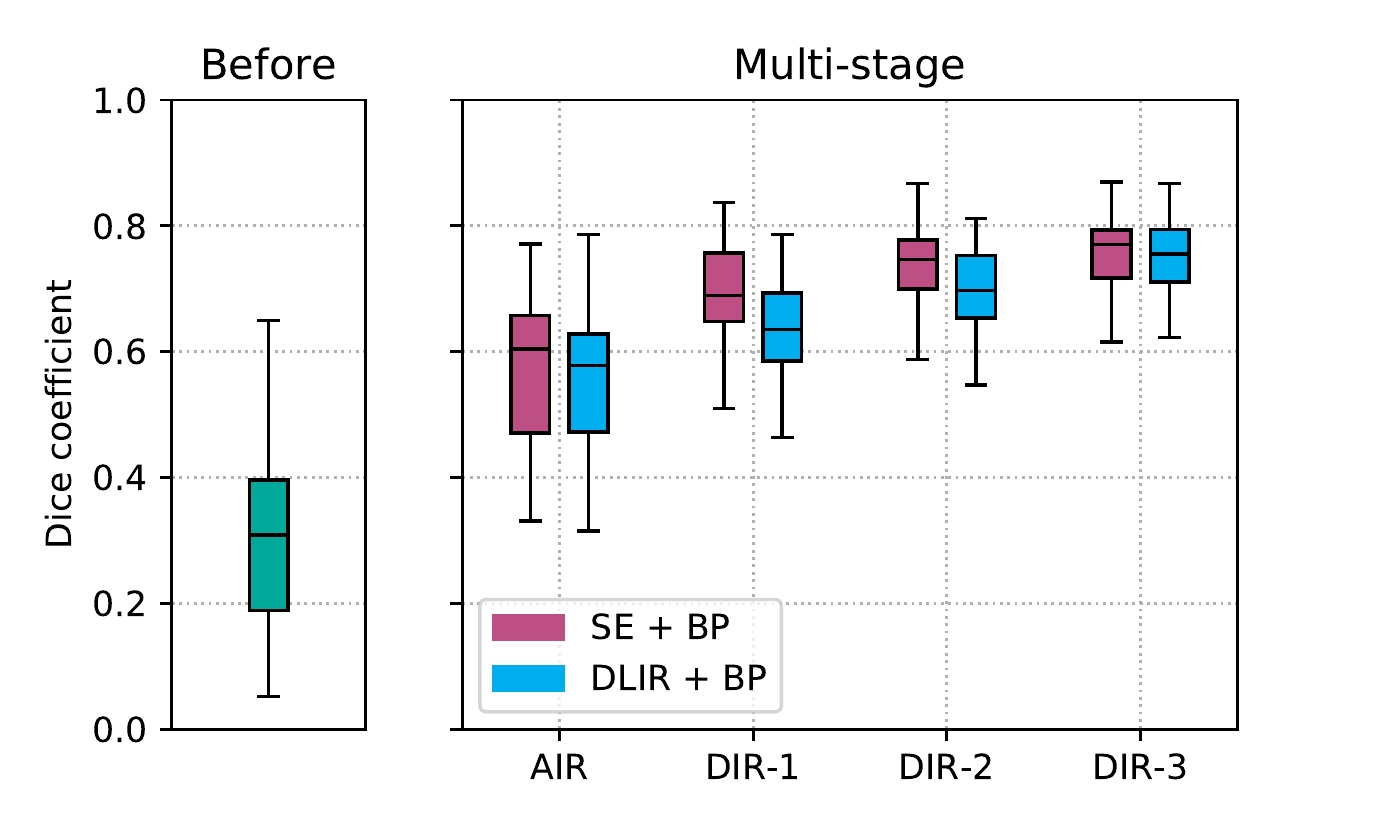}}
		\subfloat[Hausdorff distance]{\includegraphics[width = .5\textwidth]{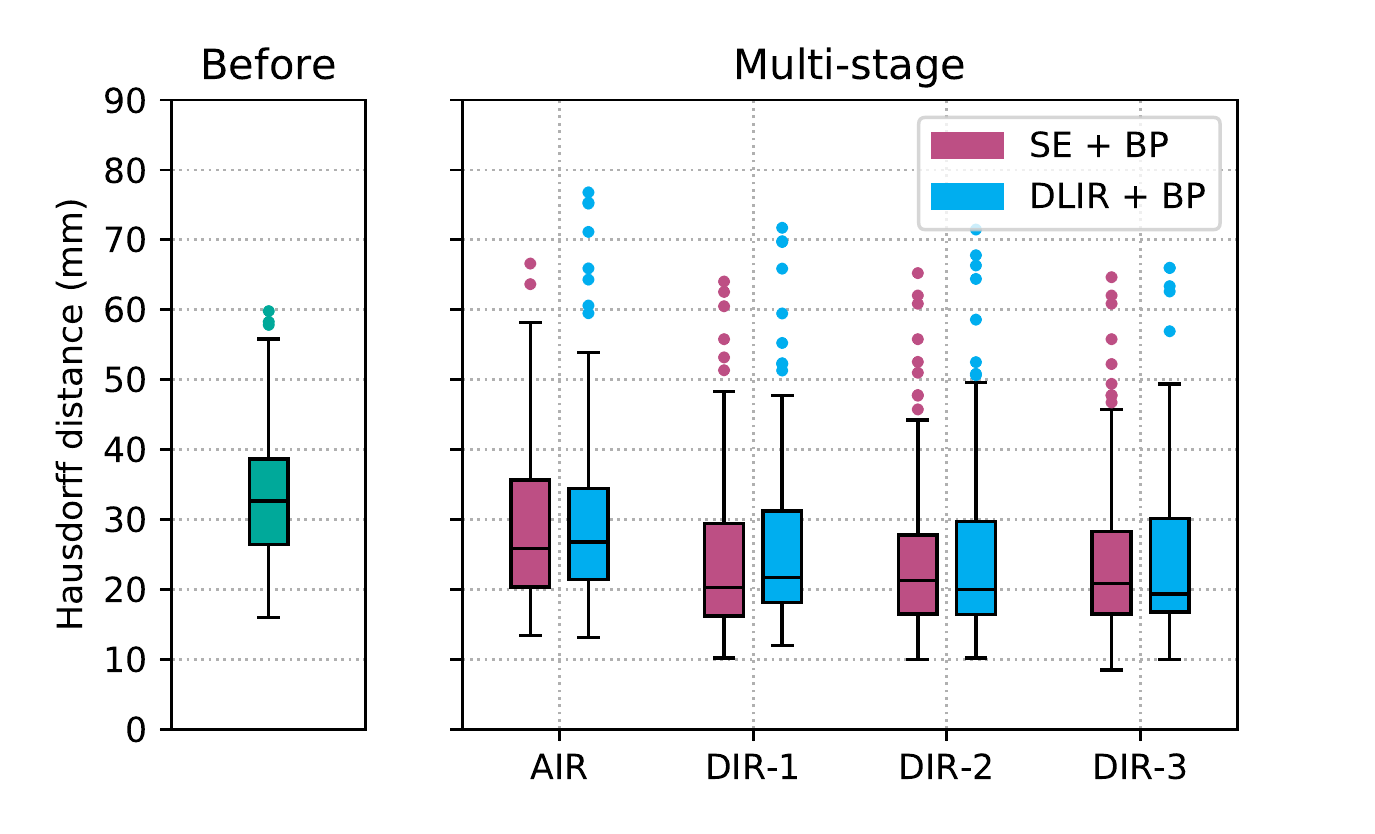}}\\
		\subfloat[Average surface distance]{\includegraphics[width =  .5\textwidth]{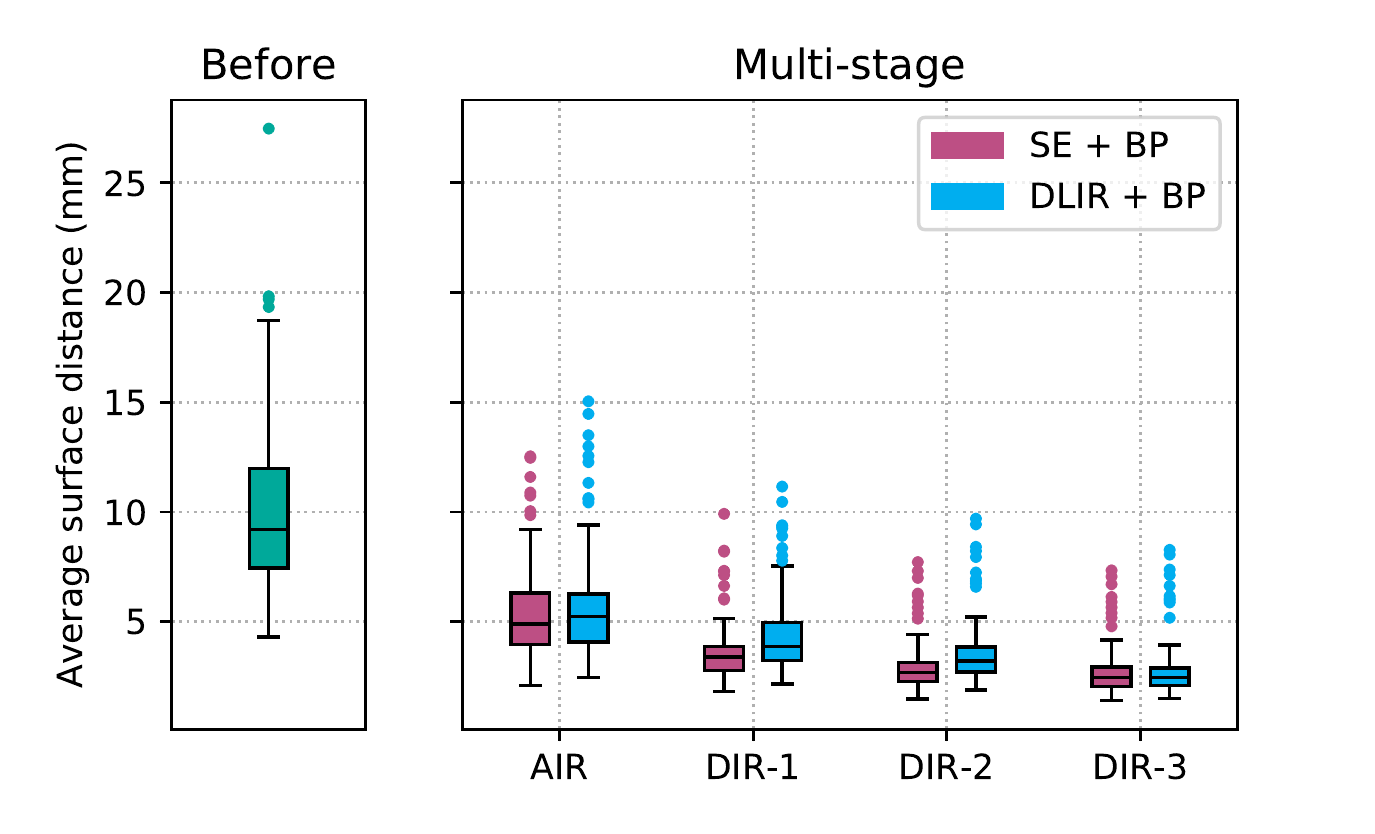}}
		\subfloat[Landmark registration error]{\includegraphics[width =  .5\textwidth]{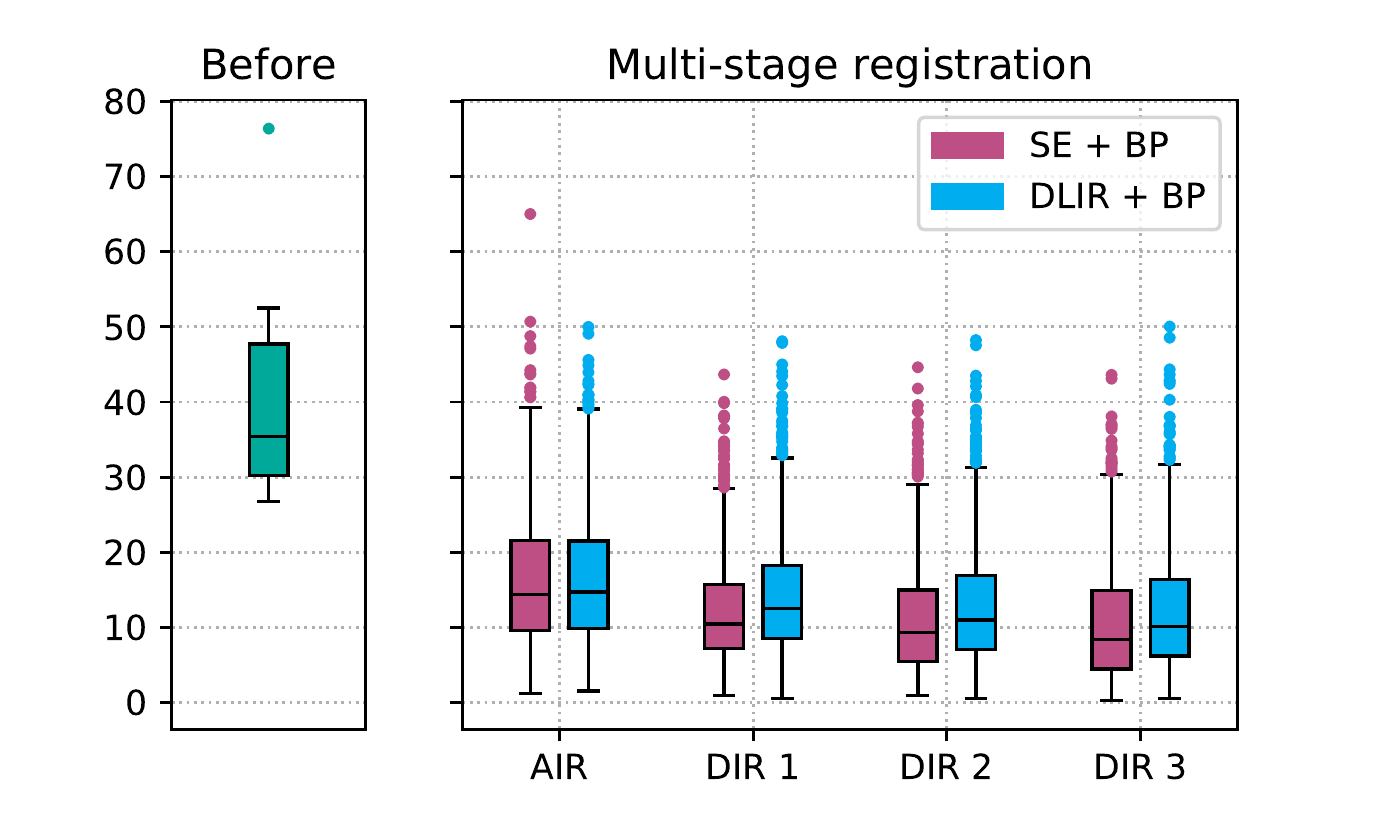}}
	\caption{Label propagation results of manual aorta delineations of inter-patient chest CT registration. Boxplots of (a) Dice, (b) Hausdorff distance, (c) average surface distance, and (d) landmark registration error are shown for conventional image registration with SimpleElastix (SE) and the DLIR framework. Left boxplots: results before image registration. Right boxplots: results of multi-stage image registration.}
	\label{fig:nlst_boxplots}
\end{figure*}

Figure~\ref{fig:nlst_boxplots} shows that Dice and ASD are similar for affine registration with conventional image registration and DLIR. The first two deformable registration stages have slightly lower Dice and higher ASD. In contrast, HD is lower for DLIR, mean that segmentations registered with DLIR have less deviations from the reference than segmentations registered with conventional image registration. In the last stage, registration performance is similar for DLIR and conventional registration, but DLIR has less outliers. The Wilcoxon signed-rank test indicated that for the final registration stage, Dice and ASD were not significantly different between conventional registration and DLIR.

Table~\ref{tab:nlst_results} gives an overview of all registration results and execution times. It shows that registration with the DLIR framework achieves quick registrations. Including image resampling, registration was took approximately 0.43\,s per image pair on a GPU.

\begin{figure}
\centering
		\includegraphics[width = .4\textwidth]{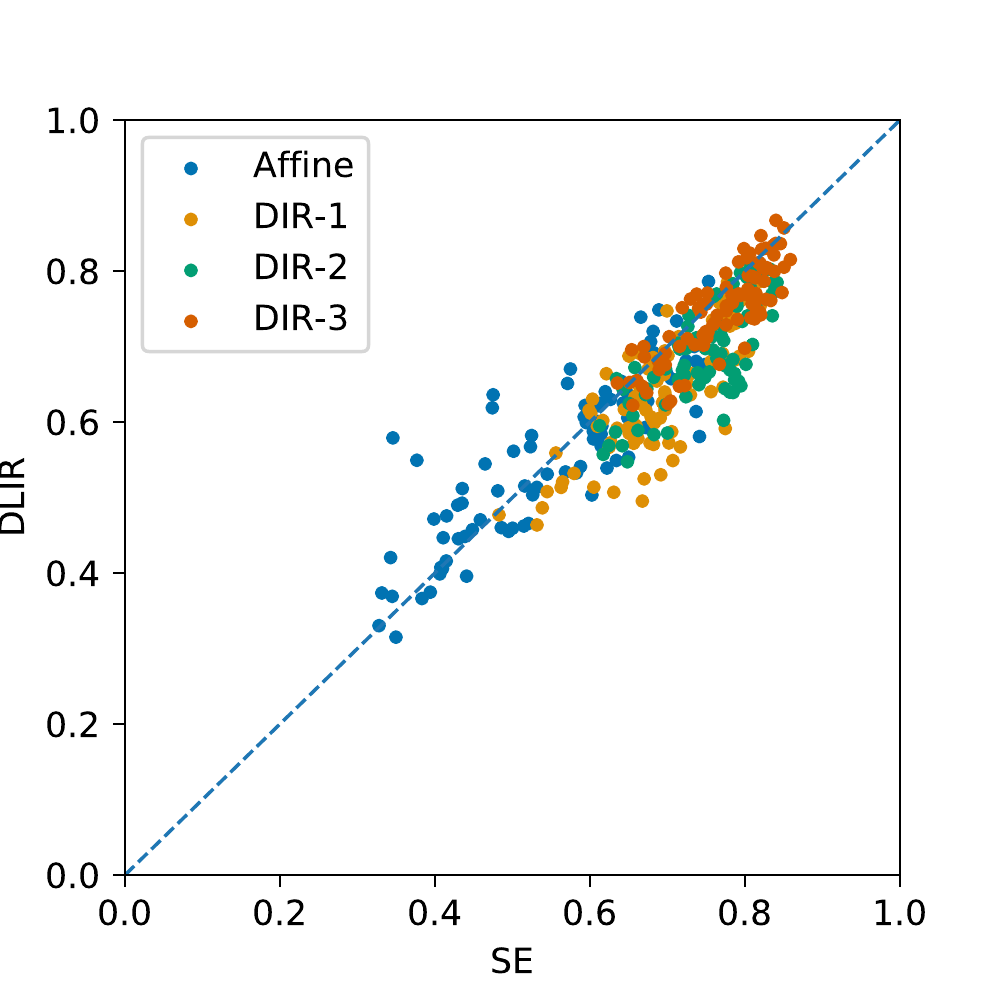}
	\caption{Scatter plot showing a comparison between Dice scores obtained with the DLIR framework and conventional intra-patient registration of chest CT. The plots show a correlation, but the dispersion of the points indicates that the registration tasks are not equally difficult for the DLIR framework and conventional registration framework.}
	\label{fig:scatterplots:nlst}
\end{figure}

Figure~\ref{fig:scatterplots:nlst}, shows that a correlation between conventional image registration and DLIR registration with respect to registration quality of image pairs. However, some registrations are more difficult for conventional image registration, while being correctly performed with DLIR, and vice versa.

\begin{table*}
\caption{Results of the inter patient chest-CT registration experiments. DLIR is compared with conventional image registration using SimpleElastix. Results are given for all stages as median $\pm$ interquartile range. Execution times are presented as mean (standard deviation) in seconds.}
\label{tab:nlst_results}
\resizebox{\textwidth}{!}{ %
\begin{tabular}{ll|ccccccc}
                              &            & Dice        & HD           & ASD         & Fraction folding (\%) & Std. dev.  Jacobian & CPU time (s) & GPU time (s)                      \\
\hline
\multicolumn{2}{l|}{Before registration}    & $ 0.31\pm0.21 $ & $ 32.62\pm12.21 $ & $ 9.21\pm4.53 $ &--       &  --     &  --   &         --                   \Tstrut\\
\hline\Tstrut
    
    \multirow{4}{*}{SE}   & AIR    & $ 0.60\pm0.19 $ & $ 25.81\pm15.34 $  & $ 4.89\pm2.36 $ & --  & --    &  $3.73(0.26)$     &  --                \\
                          & DIR-1  & $ 0.69\pm0.11 $ & $ 20.30\pm13.26 $  & $ 3.39\pm1.11 $ & $ 0.00\pm0.00 $   & $ 0.19\pm0.11 $     &        $11.67(1.07)$     & --                  \\
                          & DIR-2  & $ 0.75\pm0.08 $ & $ 21.26\pm11.31 $  & $ 2.67\pm0.87 $ & $ 0.00\pm0.08 $  & $ 0.27\pm0.13 $     & $14.83(3.37)$& -- \\ 
                          & DIR-3  & $ 0.77\pm0.08 $ & $ 20.83\pm11.81 $  & $ 2.45\pm0.89 $ & $ 0.04\pm0.19 $  & $ 0.30\pm0.15 $     &    $20.36(8.41)$ & --                            \\
\hline\Tstrut
\multirow{4}{*}{DLIR}     & AIR    & $ 0.58\pm0.16 $ & $ 26.79\pm13.05 $  & $ 5.24\pm2.19 $ & --  & --     &  $1.02(0.29)$     & $0.17(0.05)$               \\
                          & DIR-1  & $ 0.64\pm0.11 $ & $ 21.68\pm13.09 $  & $ 3.86\pm1.74 $ & $ 0.00\pm0.00 $  & $ 0.16\pm0.09 $     &  $3.85(0.99)$     & $0.18(0.05)$             \\     
                          & DIR-2  & $ 0.70\pm0.10 $ & $ 19.95\pm13.30 $  & $ 3.21\pm1.15 $ & $ 0.00\pm0.00 $  & $ 0.19\pm0.10 $     &  $8.18(2.03)$ & $0.30(0.07)$\\
                          &  DIR-3 & $ 0.75\pm0.08 $ & $ 19.34\pm13.41 $  & $ 2.46\pm0.80 $ & $ 0.75\pm1.08 $  & $ 0.45\pm0.21 $     &  $15.41(4.38)$  & $0.43(0.10)$  \\                      
\end{tabular}}
\end{table*}

\FloatBarrier

\section{Intra-Patient Registration of 4D Chest CT}
Current registration benchmark datasets unfortunately do not provide sufficient scans to train a ConvNet using the DLIR framework. Nevertheless, to give further insight in the method's performance and especially to enable reproducing our results, we performed experiments using the publicly available DIR-Lab data. The used dataset consists of ten 4D chest CTs that encompass a full breathing cycle in 10 timepoints. For each scan, 300 manually identified anatomical landmarks in the lungs in two timepoints--at maximum inspiration and maximum expiration--are provided. The landmarks serve as a reference for evaluating deformable image registration algorithms.

\subsection{ConvNet Design and Training}
Because the number of scans is very limited, we performed a leave-one-out cross-validation experiments, where one scan was used for evaluation and the nine remaining scans were used for training. The dataset size was too limited to train a ConvNet for affine registration, thus only ConvNets for deformable image registration were trained. Image intensities were clamped between {-1000} and {-200\,HU} and scaled between 0 and 1. This allowed the ConvNet to mainly focus on the anatomy of the lungs. ConvNets were trained for intra-patient registration by taking random timepoints per patient as fixed and moving images. This resulted in only 810 fixed and moving image permutations that were available for training. Ten ConvNets of similar design as used in inter-patient chest CT registration were trained in 15 hours each. Detailed experimental settings are are provided in Table~\ref{tab:dirlabtrain}. The ConvNets were trained by taking random spatially corresponding image patches of $128\times128\times64$ voxels from fixed and moving image pairs to limit memory consumption. Nevertheless, during testing, scans six to ten had to be cropped to the chest to further limit memory consumption.

\begin{table}
	\centering
	\caption{Experimental settings of the DLIR framework for training a multi-stage ConvNet for intra-patient registration of 4D chest CT from DIR-Lab data. The ConvNet consists of four deformable image registration (DIR) stages.}
	\label{tab:dirlabtrain}
	\resizebox{\columnwidth}{!}{%
	\begin{tabular}{lcccc}
		Stage & DIR-1    & DIR-2    & DIR-3 \\
		\hline\Tstrut
		Input image resolution (mm)        & $4\times4\times5$ & $2\times2\times2.5$& $1\times1\times2.5$\\
		Grid spacing (mm)        & $32\times32\times40$     & $16\times16\times20$     & $8\times8\times10$ \\
		Mini-batch size (pairs)              & $8$     & $4$      & $2$  \\
	\end{tabular}
	}
\end{table}

\subsection{Results}
\begin{figure*}
\centering
		\subfloat[]{\includegraphics[width = .4\textwidth]{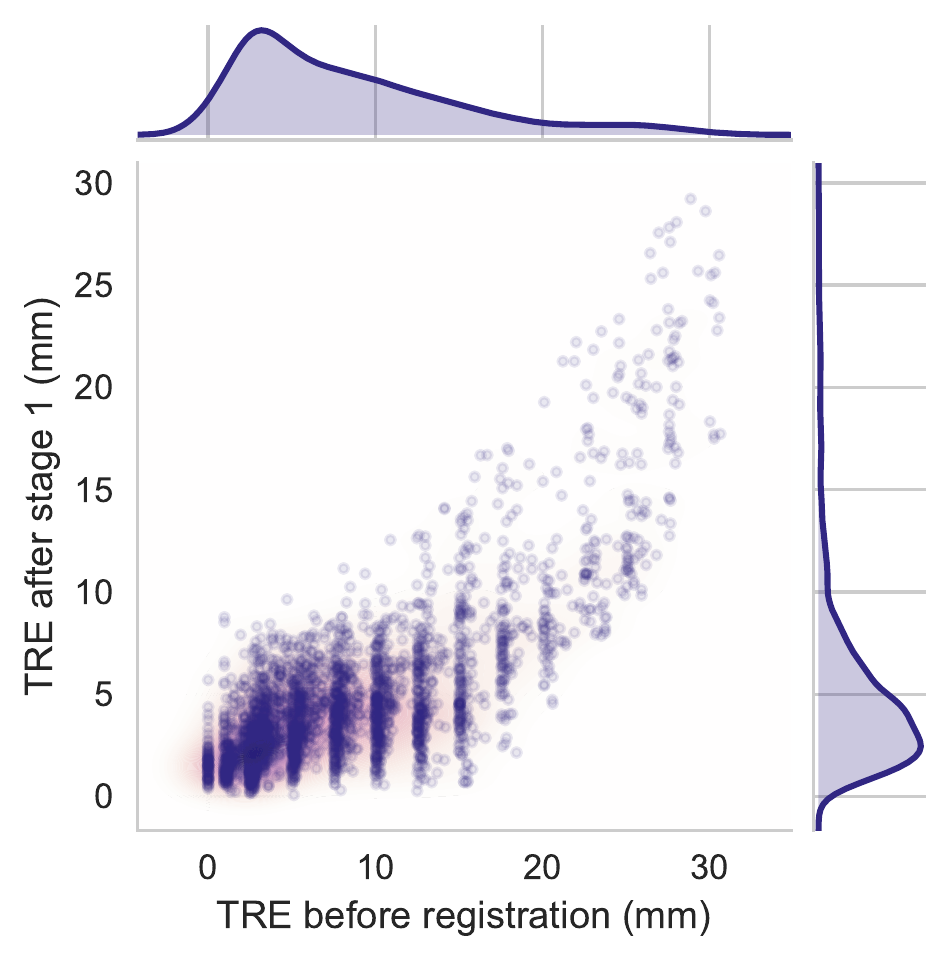}
		\label{fig:scatterplots:dirlab1}}
		\subfloat[]{\includegraphics[width = .4\textwidth]{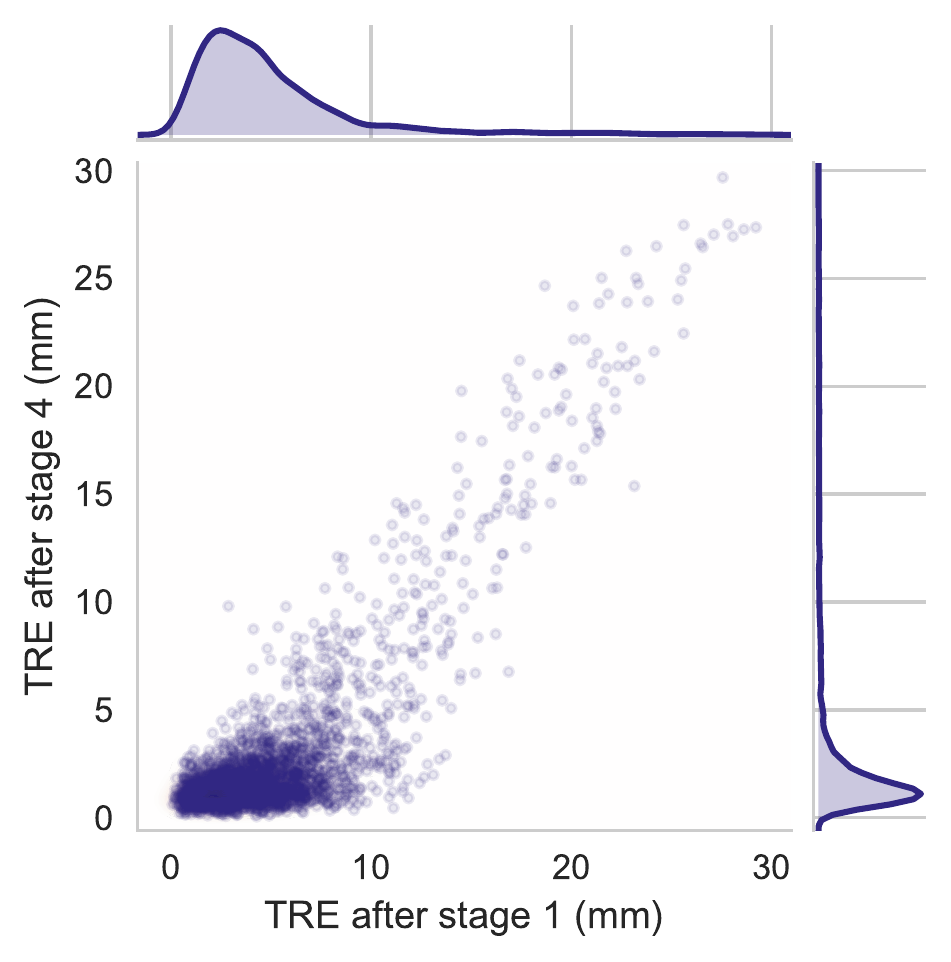}
		\label{fig:scatterplots:dirlab2}}
	\caption{Scatterplots with joint histograms illustrate that large initial deformations are underrepresented in the DIRLab dataset and that the ConvNet is unable to correctly align those with the first registration stage as shown in \protect\subref{fig:scatterplots:dirlab1}. As a consequence the ConvNet was unable to correct this in later stages as shown in \protect\subref{fig:scatterplots:dirlab2}. Nevertheless, the majority of landmarks were registered adequately by the ConvNet.}
	\label{fig:scatterplots_dirlab}
\end{figure*}

The results are listed in Table~\ref{tab:dirlab_results}, which also shows results of conventional image registration method based on Elastix \citep{berendsen2014} and a supervised deep learning based method \citep{eppenhof2018}. The final average registration error was 2.64\,mm with a standard deviation of 4.32. The error is highly influenced by outliers, likely caused by the limited dataset size. Large initial landmark distances were scarcely available for training, which influenced registration performance, as illustrated in Figure~\ref{fig:scatterplots_dirlab}. By removing 10\% of the landmarks with the largest initial registration error  more than 17.7\,mm---of which 1.47\% is coming from scan 8---an adjusted registration error is obtained of 1.63\,mm with a standard deviation of 1.67. The average registration time was 0.63\,s for multi-stage image registration, including intermediate and final image resampling.

\begin{table*}[]
\centering
\caption{Mean (standard deviation) of the registration error in mm determined on DIR-Lab 4D-CT data. From left to right: initial landmark distances (i.e. prior to registration), results of conventional image registration \citep{berendsen2014}, results of supervised deep learning method~\citep{eppenhof2018}, and registration results our proposed multi-stage DLIR. Individual registration results are shown for all ten scans.  We refer the reader to \mbox{\url{https://www.dir-lab.com/Results.html}} for a list providing results of other registration methods.}
\label{tab:dirlab_results}
\begin{tabular}{C{2.cm}|C{2.cm}|C{2.cm}|C{2.cm}|C{2.cm}C{2.cm}C{2.cm}}
     &         &  Berendsen & Eppenhof & \multicolumn{3}{c}{DLIR} \\
Scan & Initial &  et al. (2014)           & et al. (2018)         & Stage1 & Stage2 & Stage3 \\
\hline\Tstrut
Case \phantom{1}1 & $3.89 (2.78)$ & $1.00 (0.52)$ & $1.65 (0.89)$ & $2.34 (1.76)$ & $1.72 (1.37)$ & $1.27 (1.16)$ \\
Case \phantom{1}2 & $4.34 (3.90)$ & $1.02 (0.57)$ & $2.26 (1.16)$ & $2.28 (1.52)$ & $1.61 (1.31)$ & $1.20 (1.12)$ \\
Case \phantom{1}3 & $6.94 (4.05)$ & $1.14 (0.89)$ & $3.15 (1.63)$ & $3.89 (1.77)$ & $2.32 (1.58)$ & $1.48 (1.26)$ \\
Case \phantom{1}4 & $9.83 (4.85)$ & $1.46 (0.96)$ & $4.24 (2.69)$ & $3.78 (1.95)$ & $2.49 (1.90)$ & $2.09 (1.93)$ \\
Case \phantom{1}5 & $7.48 (5.50)$ & $1.61 (1.48)$ & $3.52 (2.23)$ & $3.51 (2.28)$ & $2.66 (2.15)$ & $1.95 (2.10)$ \\
Case \phantom{1}6 & $10.89 (6.96)$ & $1.42 (1.71)$ & $3.19 (1.50)$ & $7.58 (6.46)$ & $6.04 (6.64)$ & $5.16 (7.09)$ \\
Case \phantom{1}7 & $11.03 (7.42)$ & $1.49 (1.06)$ & $4.25 (2.08)$ & $5.05 (2.36)$ & $3.90 (2.46)$ & $3.05 (3.01)$ \\
Case \phantom{1}8 & $14.99 (9.00)$ & $1.62 (1.71)$ & $9.03 (5.08)$ & $8.57 (3.55)$ & $6.99 (4.52)$ & $6.48 (5.37)$ \\
Case \phantom{1}9 & $7.92 (3.97)$ & $1.30 (0.76)$ & $3.85 (1.86)$ & $6.12 (2.79)$ & $3.51 (2.02)$ & $2.10 (1.66)$ \\
Case 10 & $7.30 (6.34)$ & $1.50 (1.31)$ & $5.07 (2.31)$ & $3.76 (2.36)$ & $2.85 (2.11)$ & $2.09 (2.24)$ \\
\hline\Tstrut
Total & $8.46 (6.58)$ & $1.36 (1.01)$ & $4.02 (3.08)$ & $5.12 (4.64)$ & $3.40 (4.17)$ & $2.64 (4.32)$ \\
\end{tabular}
\end{table*}

\section{Discussion}
We have presented a new framework for unsupervised training of ConvNets for 3D image registration: the Deep Learning Image Registration (DLIR) framework. The DLIR framework exploits image similarity between fixed and moving image pairs to train a ConvNet for image registration. Labeled training data, in the form of example registrations, are not required. The DLIR framework can train ConvNets for hierarchical multi-resolution and multi-level image registration and it can achieve accurate registration results. 

Essentially the DLIR-framework can be viewed as an unsupervised training framework for STNs.
The DLIR framework shares many elements with a conventional image registration framework, as is shown in Figure~\ref{fig:framework}. In both frameworks pairs of fixed and moving images are registered by predicting transformation parameters. In both frameworks transformation parameters are inputs for a transformation model that warps the moving image. In both frameworks image similarity between the fixed and the warped moving image is used to improve transformation parameter prediction. However, while a conventional image registration framework is always used during application, the DLIR framework is only used during training of a ConvNet for image registration. After training the ConvNet can be applied for one-shot image registration of unseen images. The DLIR framework allows unsupervised training of ConvNets for affine and deformable image registration. By combining multiple ConvNets, each with its own registration task, a multi-stage ConvNet can be made that is able to perform complex registration tasks like inter-patient image registration. 

In this study three multi-stage ConvNets were trained within the DLIR framework for intra-patient registration of cardiac cine MRI, for inter-patient registration of chest CT, and for intra-patient registration of 4D chest CT. 
In all registration experiments the method showed registration results that are similar to conventional image registration but within exceptionally short execution times, which is especially desirable in time-critical applications.

The DLIR framework matched registration performance of the conventional method in intra-patient cardiac MR registration. Even though evaluation was performed with image pairs having maximum deformation between them, because evaluation pairs were taken from ES and ED time-points; while training was performed with image pairs having limited deformation between them, because training pairs were randomly taken from the full cardiac phase. Results would likely improve by training a ConvNet with a representative data-set of larger deformations, e.g. more image pairs taken from ES and ED timepoints. However, to accomplish this, the number of training scans should be substantially increased.
Likewise conclusions can be drawn for 4D chest CT registration experiments with DIR-Lab data. Performance would likely be improved when using a larger data-set with representative training data. Nevertheless, even with this very limited training set size, adequate registration results were obtained within 0.63\,s.

In inter-patient registration experiments the DLIR framework had a similar performance as the conventional image registration method in the first stages, most notably in affine registration. However the DLIR framework was slightly outperformed by the conventional method at later stages. 
This might (partially) be caused suboptimal ConvNet design choices imposed by memory limitations, e.g. the use of strided convolutions for downsampling. 
Nonetheless, the DLIR framework achieves accurate registration results with limited outliers while performing registrations faster than the conventional iterative method.

Performance of the DLIR framework is highly related to the interplay between the number of training image (or patch) pairs and registration problem complexity. For deformable registration ConvNets training is patch-based, while for affine registration ConvNets training is image based. Hence, deformable registration ConvNets allow extraction of multiple training samples from image pairs, while for affine registration ConvNets each image pair is one training sample. In intra-patient registration of cardiac MRI and inter-patient registration of chest CT the balance between number of training samples and problem complexity was adequate to match performance of conventional image registration. As expected with the DIR-Lab experiments, conventional image registration outperformed DLIR. Most likely caused by the amount of available representative training data. The employed augmentations were insufficient, and large deformations were not corrected by DLIR. Possibly by adding more training data results will improve.

Addition of a bending energy penalty mitigated occurrence of folding. In conventional image registration the penalty is used during application and as consequence it increases execution time. In DLIR the penalty is applied only during training. While it increased memory consumption and therefore in our experiments limited the number of registration stages to three, it had no effect on execution time. Yet, like in conventional image registration, full elimination of folding is not guaranteed. Nevertheless, folding was within acceptable ranges. Additional regularization during training might enforce diffeomorphism~\citep{staring2007}, with no extra cost to execution time.

The last stage of DLIR was subject to increased amounts of folding. Possibly small misregistrations of preceding stages influenced later stages, which ultimately introduced singularities. Fine-tuning the full multi-stage DLIR pipeline end-to-end might reduce this. But, owing to memory limitations, imposed by hardware and software, end-to-end training of the multi-stage ConvNets was impossible. Instead, a hierarchical training approach was used where weights of the ConvNets from preceding stages were fixed. Fixing these weights during training drastically limited memory consumption, which enabled training of large multi-stage ConvNets. Furthermore, end-to-end training of a multi-stage ConvNet could prove to be difficult: exploding gradients hampered end-to-end training in preliminary experiments using highly downsampled data. In future work stringent regularization might allow full end-to-end training of large multi-stage ConvNets, when memory issues have been dealt with.

This work employed coarse-to-fine image registration experiments such that in each registration stage maximum deformations were within the capture range of the B-spline. Given, that ConvNets were designed such that the receptive fields coincided with the B-spline capture range, the receptive fields also captured maximum deformations. In future work it would be interesting to study how DLIR would behave when dealing with deformations that are outside the receptive field and how this would affect registration of areas of uniform intensity.

The DLIR framework is able to recast conventional intensity-based image registration into a learning problem.  Thus, the framework can be extended with techniques from conventional image registration and deep learning. Features from conventional image registration, such as different transformation models like thin plate splines or direct DVF estimation can be readily implemented. 
Additionally, different image similarity metrics could be implemented; while in the current paper DLIR was employed on same-modality MR and CT data, the framework readily supports multi-modality image registration, i.e. by replacing the similarity metric for mutual information \citep{pluim2003}.
In our experiments we have used a simple ConvNet design with a limited memory footprint to demonstrate feasibility of the proposed DLIR framework. More complex ConvNet designs could be used, but complex designs are often at the cost of memory. Nevertheless, a large range of designs could be implemented in the proposed framework. In future studies we will investigate impact of other conventional image registration and deep learning techniques on image registration robustness and accuracy.

\section{Conclusion}
We presented the Deep Learning Image Registration framework for unsupervised affine and deformable image registration with convolutional neural networks. We demonstrated that the DLIR framework is able train ConvNets without training examples for accurate affine and deformable image registration within very short execution times.

\section*{Acknowledgment}
This work is part of the research programme ImaGene with project number 12726, which is partly financed by the Netherlands Organisation for Scientific Research (NWO). 

The authors thank the National Cancer Institute for access to NCI's data collected by the National Lung Screening Trial. The statements contained herein are solely those of the authors and do not represent or imply concurrence or endorsement by NCI.

\section*{References}

\bibliography{bibliography}

\end{document}